\documentclass{article}

\usepackage[preprint]{neurips_2026}

\usepackage[utf8]{inputenc} % allow utf-8 input
\usepackage[T1]{fontenc}    % use 8-bit T1 fonts
\usepackage{hyperref}       % hyperlinks
\usepackage{url}            % simple URL typesetting
\usepackage{booktabs}       % professional-quality tables
\usepackage{amsfonts}  
\usepackage[linesnumbered,ruled,vlined]{algorithm2e}% blackboard math symbols
\usepackage{nicefrac}       % compact symbols for 1/2, etc.
\usepackage{microtype}      % microtypography
\usepackage{xcolor}         % colors
\usepackage{times}
\usepackage{soul}
\usepackage{graphicx}
\usepackage{bbding}
\usepackage{amsmath}
\usepackage{amsthm}
\usepackage{amssymb}
\usepackage{algorithmic}
\usepackage{subfigure}
\usepackage{multirow}
\usepackage{wrapfig}
\usepackage{colortbl}

\newtheorem{definition}{Definition}

\PassOptionsToPackage{number,compress}{natbib}
\title{Navigating by Old Maps: The Pitfalls of Static Mechanistic Localization in LLM Post-Training}

\author{%
  Hang Chen \\
  School of Computer Science and Technology\\
 Xi'an Jiaotong University\\
  \texttt{albert2123@stu.xjtu.edu.cn} \\
  \And
   Jiaying Zhu\\
  School of Computer Science and Engineering\\
 The Chinese University of Hong Kong\\
  \texttt{jyzhu24@cse.cuhk.edu.hk} \\
  \And
   Hongyang Chen\\
  Shaanxi Co., Ltd(Xi'an 710077, China)\\
 China Mobile Group\\
  \texttt{chenhongyang@sn.chinamobile.com} \\
  \And 
  Hongxu Liu\\
  College of Computing and Data Science\\
 Nanyang Technological University\\
  \texttt{hongxu001@e.ntu.edu.sg} \\
  \AND
   Xinyu Yang\thanks{Corresponding author}\\
  School of Computer Science and Technology\\
 Xi'an Jiaotong University\\
  \texttt{yxyphd@mail.xjtu.edu.cn} \\
  \And
  Wenya Wang\thanks{Corresponding author}\\
  College of Computing and Data Science\\
 Nanyang Technological University\\
  \texttt{wangwy@ntu.edu.sg} \\
}

\begin{document}

\maketitle

\begin{abstract}
  The "Locate-then-Update" paradigm has become a predominant approach in the post-training of large language models (LLMs), identifying critical components via mechanistic interpretability for targeted parameter updates. However, this paradigm rests on a fundamental yet unverified assumption: can mechanisms derived from current static parameters reliably guide future dynamic parameter updates? To investigate this, we systematically track the structural evolution of Transformer circuits throughout the supervised fine-tuning (SFT) process, revealing the underlying dynamics of task mechanisms. We introduce three novel metrics—Circuit Distance, Circuit Stability, and Circuit Conflict—to analyze circuit evolution across three dimensions: neural migration, semantic stability, and cross-task interference. Our empirical results reveal that circuits inherently exhibit "Free Evolution" during parameter updates. Consequently, static mechanisms extracted from current states inevitably suffer from temporal latency, making them fundamentally inadequate for guiding future states. Moreover, by deconstructing the "illusion of effectiveness" in existing methods, this work underscores the necessity of "foresight" in mechanistic localization and proposes a predictive framework for future research. Our code is available at \url{https://github.com/Zodiark-ch/MechLocalization}. 
\end{abstract}

\section{Introduction}
Post-training optimization of large language models (LLMs) refers to the targeted refinement of pre-trained language models equipped with powerful general capabilities~\citep{lai2025survey,xiao2023smoothquant}. Strategies such as supervised fine-tuning (SFT)~\citep{hu2022lora}, reinforcement learning~\citep{havrilla2024teaching}, parameter editing~\citep{yao2023editing}, or vector steering~\citep{cao2024personalized} are employed to marginally alter model parameters. This ensures the model better aligns with practical application scenarios while preserving its general capabilities. From an optimization perspective, post-training processing aims to achieve optimal performance on a new \textbf{target task} while maintaining existing capabilities (hereafter referred to as \textbf{pervasiveness tasks})~\citep{zhang2026locate}.

To mitigate catastrophic forgetting on pervasiveness tasks, recent studies increasingly adopt a "\textbf{locate-then-update}" paradigm across applications like model unlearning~\citep{wu2023depn,li2025effective}, knowledge editing~\citep{meng2022locating,dai2022knowledge}, and reinforcement learning\citep{yan2026spurious}. This paradigm relies on \textbf{Mechanistic Localization}—using mechanistic interpretability to identify the minimal parameter space responsible for the target skill. Post-training parameter updates are then exclusively confined to this localized region.

However, recent studies show that Mechanistic Localization often lacks completeness~\citep{chen2025rethinking} and exclusiveness~\citep{hase2023does}, raising a critical question: acts it merely as a ``placebo''? Specifically, \textbf{can a static snapshot of mechanistic interpretability findings genuinely guide the dynamic process of future parameter updates?} As illustrated in Figure~\ref{figintro}, if full-parameter SFT causes the target task's critical components to shift from $B_1, B_2$ to $A_1, A_2$, localization based solely on pre-update parameters will prematurely freeze $A_1$ and $A_2$.  Therefore, it remains unclear whether Mechanistic Localization genuinely prevents conflicts or improperly constrains the target mechanism's natural evolution.
To address this, we break down this problem into two specific research questions:

\begin{wrapfigure}{r}{0.6\linewidth}
\vspace{-4mm}
  \begin{center}
    \includegraphics[width=\linewidth]{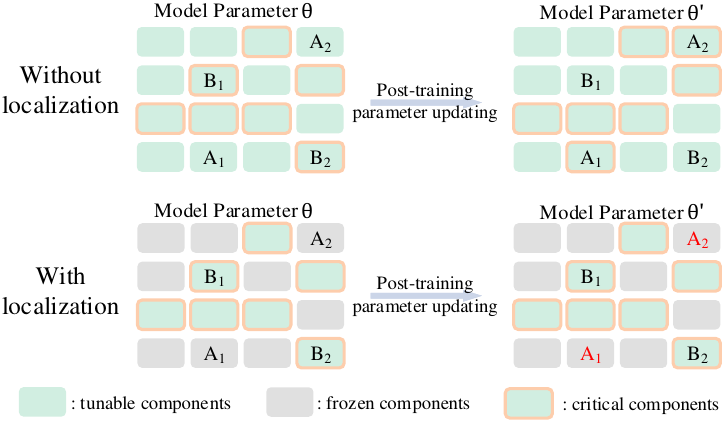}
  \end{center}
  \vspace{-4mm}
  \caption{Differences in mechanism localization in post-training SFT with and without localization}
  \vspace{-4mm}
  \label{figintro}
\end{wrapfigure}
\begin{itemize}
    \item \textbf{RQ1}: Without localization, do the critical components of the target skill change during parameter updates? If so, how do they evolve?
    \item \textbf{RQ2}: Does post-training with localization truly improve performance on the target task and mitigate conflicts with pervasiveness tasks?
\end{itemize}

In this paper, we employ \textbf{SFT} as the representative of post-training process to intuitively observe the evolutionary process, defining the \textbf{component}\footnote{Specifically, independent parameter matrices like $W_q, W_k, W_v, W_o$ in attention heads and $W_{\text{up}}, W_{\text{down}}$ in MLPs.} as the minimal update unit. For Mechanistic Localization, we adopt \textbf{circuit discovery}~\citep{conmy2023towards,syed2024attribution}, which comprehensively captures overarching mechanisms and inter-component connections. To address \textbf{RQ1}, we introduce \textbf{circuit distance} to quantify critical component migration, and \textbf{circuit stability} to evaluate the model's mastery of the task mechanism. For \textbf{RQ2}, alongside direct performance observation, we propose \textbf{circuit conflict} to measure how effectively localization prevents mechanistic clashes between the target and pervasiveness tasks. Through extensive experiments, we draw the following conclusions:
\begin{itemize}
    \item \textbf{Divergent Free Evolution:} Without localization, critical components evolve freely, exhibiting distinct structural patterns: attention mechanisms undergo drastic shifts, whereas MLP components remain relatively stable.
    \item \textbf{Temporal Lag of Static Localization:} Because circuits inherently exhibit free evolution, utilizing current parameter states as a reference for future parameters suffers from severe latency and temporal lag.
    \item \textbf\textbf{The Illusion of Effectiveness:} The perceived success of existing Mechanistic Localization methods relies heavily on their application to MLP-dominated, knowledge-centric downstream tasks(e.g., knowledge editing, unlearning). 
\end{itemize}

Ultimately, this paper reveals that Mechanistic Localization exhibits a critical \textbf{lag} during dynamic updates. To better optimize target performance and minimize mechanistic conflicts, we explore the necessity of more advanced, dynamic localization paradigms in Section~\ref{sec:discussion}.

\section{Preliminaries}
We denote a well-trained LLM as $\mathcal{M}=\langle\mathcal{G}, \theta\rangle$, where $\theta$ represents the states of all trainable parameters. The computational graph $\mathcal{G}=\langle\mathcal{V}, \mathcal{E}\rangle$ models the forward pass, with $\mathcal{V}$ comprising all components (i.e., parameter matrices such as $W_q, W_k, W_v, W_o, W_{\text{up}}, W_{\text{down}}$) and $\mathcal{E}$ denoting their activation connections (e.g., $W_o \rightarrow W_{\text{up}}$).

\subsection{Post-Training Processing}
We define post-training processing as modifying a target task's mechanism while preserving pre-existing capabilities (pervasiveness tasks), with typical applications including model unlearning~\citep{liu2025rethinking} and knowledge editing~\citep{wang2024knowledge}. To intuitively observe intermediate dynamics, we formalize post-training processing as a multi-objective fine-tuning task\footnote{We demonstrate in Appendix~\ref{suppmorefuture} that our conclusions also hold for non-continuous post-training methods like knowledge editing.}. Given initial parameters $\theta$, a target dataset $\mathcal{D}_{t}$ which requires input $x$ yields an output $y_t$, and a pervasiveness dataset $\mathcal{D}_{p}$, any input $x$ yields an output $y$ that follows $p(y|x,\theta)$. The updated parameters $\theta '$ are optimized via:
\begin{equation}
    \min_{\theta '} \mathbb{E}_{(x, y_t)\in \mathcal{D}_{t}}[\mathcal{L}(y_t|x;\theta ')]+\lambda \mathbb{E}_{(x, y)\in \mathcal{D}_{p}}[\mathcal{L}(y|x;\theta ')]
\end{equation}
where $\lambda \geq 0$ is the regularization parameter. Essentially, this objective ensures $\theta '$ adapts to the target behavior specified by $\mathcal{D}_{t}$ while leaving the forward pass for task-irrelevant inputs unaltered.

\subsection{Circuit Discovery}
We employ circuit discovery as our mechanistic interpretability technique. Compared to alternatives, it better deconfounds correlations (vs. gradient/magnitude methods~\citep{li2016visualizing,tang2024language}), captures holistic mechanisms (vs. causal interventions~\citep{stolfo2023mechanistic}), and provides stronger theoretical foundations (vs. probing~\citep{ju2024large} or vocabulary lenses~\citep{belrose2023eliciting}). It seeks to identify a minimal subgraph (circuit) $\mathcal{C} \subset \mathcal{G}$ capturing the task-relevant behavior of a target dataset~\citep{elhage2021mathematical,conmy2023towards,rai2024practical}, optimized via:
\begin{equation}
    \arg \min_{\mathcal{C}}\mathbb{E}_{(x)\in \mathcal{D}_{t}}[D(p_{\mathcal{G}}(y|x)||p_{\mathcal{C}}(y|x))], ~~s.t.~1-|\mathcal{C}|/|\mathcal{G}|\geq s
    \label{eqtnosing}
\end{equation}
where $s$ is the sparsity constraint and $D$ measures the output divergence between $\mathcal{G}$ and $\mathcal{C}$. The nodes and edges within $\mathcal{C}$ are thus deemed the most critical components for processing $\mathcal{D}_{t}$\footnote{Note that the discovery target dataset $\mathcal{D}_{t}$ can correspond to either the post-training target task or a pervasiveness task, depending on which mechanism is being observed.}.

\subsection{Locate-then-Update Paradigm}\label{seclocatethenupdate}
In summary, the locate-then-update paradigm consists of two steps. \textbf{First} (\textbf{locating}), mechanistic interpretability (here, circuit discovery) identifies the critical component set $\mathcal{C}=\langle\mathcal{V}_t, \mathcal{E}_t\rangle$ for $\mathcal{D}_{t}$ (We show the details about logical circuit discovery in Appendix~\ref{supplogicalcircuit}). \textbf{Second} (\textbf{updating}), the remaining components $\mathcal{V}^*=\mathcal{V} \setminus \mathcal{V}_t$ are frozen, followed by target-specific post-training on $\mathcal{D}_{t}$. Recent literature has advanced both phases: localization improvements include jointly considering $\mathcal{D}_{t}$ and $\mathcal{D}_{p}$~\citep{jia2024wagle}, ensembling multiple interpretability methods~\citep{li2025effective}, and utilizing low-dimensional projections~\citep{muhamed2025saes}; meanwhile, updating enhancements employ diverse fine-tuning strategies such as gradient ascent~\citep{liu2022continual}, direct preference optimization~\citep{mainitofu}, and negative preference optimization~\citep{zhangnegative}.

\section{Evaluation Metrics}
This section introduces three metrics (summarized in Table~\ref{tabmetrics}) to evaluate mechanism evolution. To address \textbf{RQ1}, \textbf{Circuit Distance} ($CD$) and \textbf{Circuit Stability} ($CS$) evaluate a single mechanism's evolution: given a target task's logical circuit (Appendix~\ref{supplogicalcircuit}), $CD$ measures component migration, while $CS$ assesses knowledge consolidation. To address \textbf{RQ2}, we propose \textbf{Circuit Conflict} ($CC$) to quantify inter-mechanism interference. Combined with intrinsic task performance metrics - which gauge capability enhancement or retention - $CC$ comprehensively evaluates multi-task interactions within the Locate-then-Update paradigm.

\begin{table}[ht]
  \caption{An overview of the three circuit metrics.}
  \label{tabmetrics}
  \centering
  \resizebox{0.85\textwidth}{!}{
  \begin{tabular}{llll}
    \toprule
    \textbf{Metrics} & \textbf{Object} & \textbf{Data Format} & \textbf{Function} \\
    \midrule
    $CD$ & Single circuit & Circuit Graph & How do components transfer? \\
    $CS$ & Single circuit & Circuit Distribution & How well is the mechanism learned? \\
    $CC$ & Multiple circuits & CNF from Circuits & How do mechanisms conflict? \\
    \bottomrule
  \end{tabular}}
\end{table}

\subsection{Circuit Distance ($CD$)}
In circuit discovery, an edge $W_i \rightarrow W_j \in \mathcal{C}$ is identified by measuring the output variance under causal interventions. An edge is retained if its causal effect $I(W_i \rightarrow W_j)$ exceeds a threshold $\tau$:
\begin{equation}
    I(W_i \rightarrow W_j)=\left|\mathbb{L}(x|\text{do}(W_i \rightarrow W_j))-\mathbb{L}(x)\right|> \tau
\end{equation}
where $\mathbb{L}$ denotes the output logits, and $\text{do}(\cdot)$ represents activation patching. For these interventions, we default to interchange ablation~\citep{heimersheim2024use,vig2020investigating,chan2022causal,goldowsky2023localizing}.

To quantify the migration of critical pathways across parameter states $\theta$ and $\theta'$, we employ the Manhattan distance over the computational graph $\mathcal{G}=\langle\mathcal{V}, \mathcal{E}\rangle$.  Specifically, for a computational graph $\mathcal{G}=\langle\mathcal{V}, \mathcal{E}\rangle$ with $N$ components, we associate each edge in $\mathcal{E}$ with its causal effect, denoted as $\mathcal{E}=\{W_i \rightarrow W_j, I(W_i \rightarrow W_j)\}$. This approach captures continuous variations more effectively than discrete metrics (e.g., Hamming distance). The Circuit Distance ($CD$) is defined as:
\begin{equation}
    CD=D_{\mathcal{G}}(\mathcal{G}^{\theta}, \mathcal{G}^{\theta '})=\sum_{(W_i,W_j \in \mathcal{E})}\left| I(W_i \rightarrow W_j)^{\theta}-I(W_i \rightarrow W_j)^{\theta '}\right|
\end{equation}
To account for varying logit baselines across tasks, we normalize $CD$ using the maximum empirical range of $I(\cdot)$. Ultimately, $CD$ reflects the extent of mechanism transformation by aggregating the absolute shifts in causal effects across all components.

\subsection{Circuit Stability ($CS$)}

Beyond tracking mechanism migration, it is crucial to assess the model's mastery of the mechanism. Recent studies~\citep{sun2025circuit} indicate that if the learning of the target mechanism has not converged, different sampled subsets of $\mathcal{D}_{t}$ will yield significantly divergent circuits. We quantify this using \textbf{Circuit Stability} ($CS$), defined as the expected Spearman's rank correlation ($\rho$) between circuits derived from two i.i.d. subsets $s, s'$ (each 20\% of $\mathcal{D}_t$):
\begin{equation}
    CS(\theta ')=\mathbb{E}_{s, s' \in \mathcal{D}_{t}}[\rho(\mathcal{C}^{\theta '}_{s}, \mathcal{C}^{\theta '}_{s '})]
\end{equation}

Here, $\rho$ ranks the causal effects of all edges. Specifically, we take circuits $\mathcal{C}_{s}$ and $\mathcal{C}_{s '}$, and rank $\mathcal{E}=\{W_i \rightarrow W_j, I(W_i \rightarrow W_j)\}^{N}_{(i,j)}$ based on $\mathcal{C}_{s}(\mathcal{E})$ and $\mathcal{C}_{s '}(\mathcal{E})$, respectively.
Therefore, $CS$ reflects the stability trend of the target mechanism across two parameter states by measuring the correlation between circuits derived from different randomly sampled subsets. A large $|CS|$ indicates high cross-sample consistency, meaning knowledge is effectively consolidated. Conversely, a small $|CS|$ implies the information remains scattered and unsettled, undergoing further refinement. Combining $CD$ and $CS$ yields four joint states:

\begin{itemize}
    \item \textbf{Large $CD$, Large $|CS|$}: The model preserves the firmly consolidated information of the mechanism but actively migrates its storage to a new set of components.
    \item \textbf{Small $CD$, Small $|CS|$}: The model continues to utilize the original components but actively updates and refines the unsettled internal information within them.
    \item \textbf{Large $CD$, Small $|CS|$}: The model undergoes a reorganization, simultaneously updating the unsettled internal information and migrating its storage to a new set of components.
    \item \textbf{Small $CD$, Large $|CS|$}: The model maintains a highly stable state, retaining both the consolidated information of the mechanism and the original components storing it.
\end{itemize}

An intuitive, albeit imprecise, metaphor is that $CD$ determines whether to change the \textit{glass} holding the cocktail, while $CS$ determines whether to refine the cocktail's \textit{recipe}. $CD$ and $CS$ not only reflect the evolutionary state of a single mechanism but also serve as a ``baseline evolution'' for studying multiple mechanisms. By comparing against this baseline, we can better analyze the practical impact of localization on multiple mechanisms.

\subsection{Circuit Conflict ($CC$)}

% Existing research~\citep{chen2025clue} indicates that when conflicting components between circuits of different mechanisms significantly decrease, the corresponding tasks tend to perform better during SF. We define conflicting components as critical components that must simultaneously satisfy the optimization objectives of multiple tasks. For instance, if component $W_i$ is required for optimizing both the target task and a pervasiveness task, it is highly challenging for $W_i$ to achieve optimality for multiple objectives simultaneously. From the perspective of component polysemanticity, this occurs because $W_i$ encodes essential capabilities required by both tasks. Therefore, during the evolution of post-training processing, we desire a decreasing trend in the number of conflicting components to increase the likelihood of achieving superior independent performance on each task.

We define \textbf{conflicting components} as critical components shared across multiple optimization targets' circuits. Reducing these components typically enhances post-training performance~\citep{chen2025clue}, as sharing induces optimization competition rather than synergy. For instance, a polysemantic component $W_i$ encoding both ``sports'' and ``geography'' faces competing gradient requirements during multi-objective optimization within ``sports'' and ``geography'' tasks, preventing simultaneous optimality. Conversely, disentangling these semantics into distinct components $W_m$ and $W_n$ enables independent tuning to task-specific peaks. Therefore, a decreasing trend in conflicting components during evolution is highly desirable, signifying a shift toward specialized neural representations that facilitate superior independent performance.

However, identifying conflicting components by simply intersecting circuits yields biased conclusions due to component non-exclusivity (e.g., redundant components effectively forming an ``OR gate''). To resolve this, we leverage logical circuits~\citep{chen2025clue} to map task circuits into Conjunctive Normal Form (CNF) clauses, framing conflict detection as a Boolean satisfiability (SAT) problem. Let $\Phi_{t}$ and $\Phi_{p}$ denote the CNFs for the target and pervasiveness circuits, respectively. We define Circuit Conflict ($CC$) as:
\begin{equation}
    CC(\theta)=\min_{n}{\text{UNSAT}(\Phi^{\theta}_{t} \land \Phi^{\theta}_{p})}
\end{equation}
where $\text{UNSAT}$ computes the \textbf{UNSAT Core}---the minimum number of conflicting clauses---using a SAT solver~\citep{selsam2019guiding,cimatti2007simple,d2010alloy+}. This formulation naturally expands for multiple pervasiveness tasks (e.g., $\Phi_{t} \land \Phi_{p1} \land \dots$).

By extracting the UNSAT Core across parameter states, $CC$ strictly quantifies the evolutionary trend of conflicts. A larger $CC$ signifies intensified interference from pervasiveness tasks, which inherently hinders optimal target performance. Ultimately, $CC$ bypasses superficial performance fluctuations, directly revealing whether multiple mechanisms evolve synergistically.

\section{Experiments}
\label{sec:experiments}

Our evaluation framework is grounded in a Supervised Fine-Tuning (SFT) pipeline. The primary optimization objective is to master a target task while preserving performance on a pervasiveness task. We analyze the underlying mechanistic dynamics by evaluating task circuits at each SFT step (extraction details in Appendix~\ref{supplogicalcircuit}). 

Building upon this, we structure our experiments around two distinct SFT pipelines:
\begin{itemize}
    \item \textbf{Free Evolution:} Without Mechanistic Localization, we fine-tune all parameters to investigate whether, how, to what extent, and why circuits change during parameter evolution.
    \item \textbf{Localization Evolution:} Adhering to the ``locate-then-update'' paradigm (Section~\ref{seclocatethenupdate}), we identify critical components (Appendix~\ref{supplogicalcircuit}), freeze the others, and perform SFT. By comparing this against free evolution, we explore whether localization genuinely benefits parameter updates and investigate why existing methods ostensibly improve task performance.
\end{itemize}

We employ Mistral-7B and LLaMA3-8B\footnote{Mistral-7B: \url{https://huggingface.co/mistralai/Mistral-7B-v0.1}. Llama3-8B: \url{https://huggingface.co/meta-llama/Meta-Llama-3-8B}} as baseline models. For target and pervasiveness tasks, we select 15 datasets widely adopted in mechanistic interpretability and post-training research: OpenBookQA~\citep{OpenBookQA2018}, Gender~\citep{mathwin2023identifying}, RTE~\citep{dagan2022recognizing}, IOI~\citep{wanginterpretability}, Docstring~\citep{heimersheim2023circuit}, SST2~\citep{socher-etal-2013-recursive}, Winogrande~\citep{ai2:winogrande}, Reverse~\citep{lindner2023tracr}, Greater Than~\citep{hanna2023does}, FEVER~\citep{thorne-etal-2018-fever}, zsRE~\citep{levy-etal-2017-zero}, Induction~\citep{conmy2023towards}, Bool~\citep{suzgun2023challenging}, Arithmetic~\citep{brown2020language}, and SA~\citep{yu2024functional}. We track SFT evolution across $10$ epochs, subdividing each into $20$ observation steps to capture granular dynamics. Comprehensive setups are detailed in Appendix~\ref{suppexdetail}. The performance fluctuations of all SFT experiments across $5$ random seeds are tightly bound within a narrow $95\%$ confidence interval (t=2.776) and we report the representative outcomes for clarity.

\subsection{Free Evolution}
\label{subsec:free_evolution}
In this subsection, we focus on \textbf{RQ1} (\textit{Without localization, do the critical components of the target skill change during parameter updates? If so, how do they evolve?}). Specifically, we decompose it into three sub-questions:
\begin{itemize}
    \item \textbf{RQ1-a}: Does the task mechanism evolve during parameter updates without localization?
    \item \textbf{RQ1-b}: If evolution exists, what are the underlying patterns of this evolution?
    \item \textbf{RQ1-c}: What factors influence this evolution, and how do they exert their effects?
\end{itemize}

\begin{figure*}
    \centering
    \subfigure[Circuit Distance]{
    \includegraphics[width=0.47\linewidth]{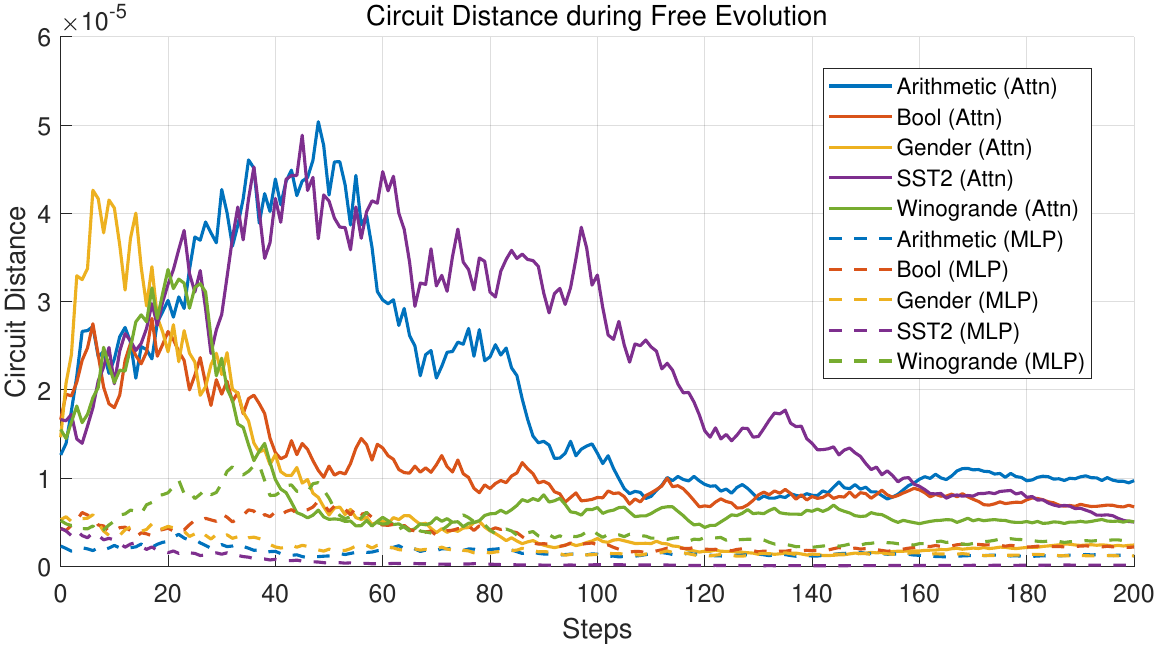}}
    \subfigure[Circuit Stability]{
    \includegraphics[width=0.47\linewidth]{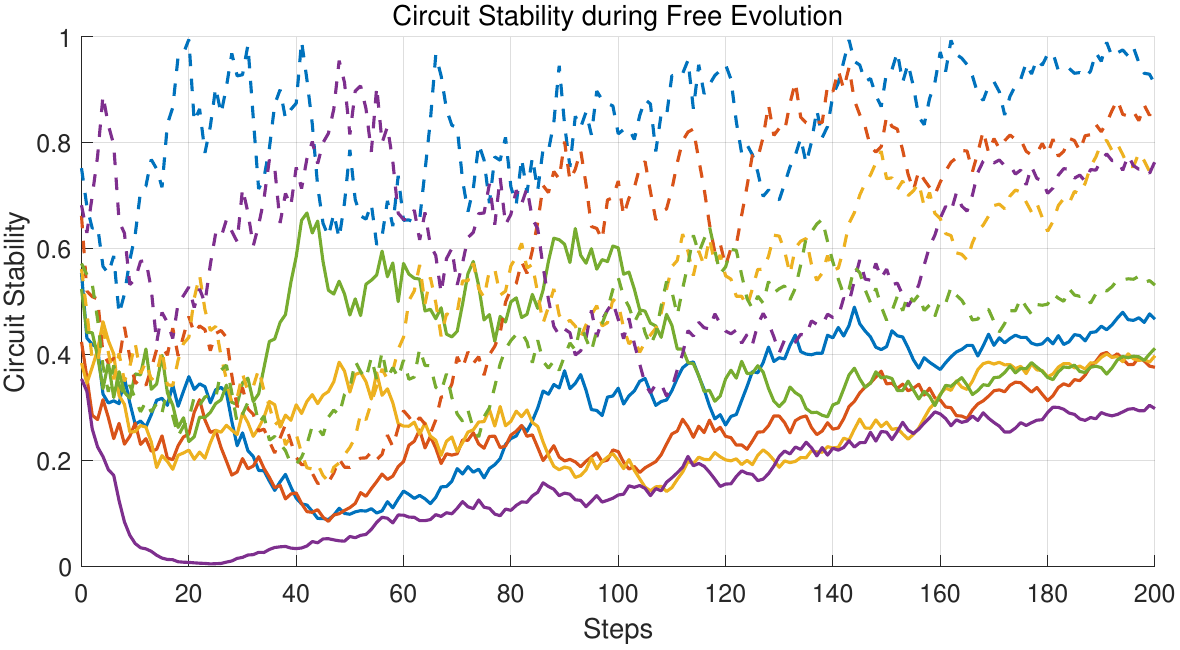}}
    \caption{line plots of different target tasks on the Mistral-7B model in terms of Circuit Distance ($CD$) and Circuit Stability ($CS$).}
    \label{figfreeevomistral}
\end{figure*}

To address \textbf{RQ1-a} and \textbf{RQ1-b}, we evaluate mechanism evolution on Mistral-7B and LLaMA3-8B across five target tasks (Arithmetic, SST2, Winogrande, Bool, Gender) against a general pervasiveness task (OpenBookQA). Circuits are extracted via the Logical Circuit Framework with EdgePruning (Appendix~\ref{suppedgepruning}). Figure~\ref{figfreeevomistral} plots the $CD$ and $CS$ trajectories during SFT on Mistral-7B, where $CD$ tracks the shift from the initial parameters ($\theta^{0}$) to the current step ($\theta^{s}$). Full results, including LLaMA3-8B findings and additional task performance metrics, are detailed in Appendix~\ref{suppfreeevo}.

These results definitively answer \textbf{RQ1-a}: without localization, task mechanisms spontaneously shift during parameter updates, driving substantial circuit migration. Notably, even well-mastered tasks like Gender (initial accuracy $>85\%$) undergo drastic migration, replacing numerous critical components (Table~\ref{tab:app_mastery_level}). Ablation studies (Tables~\ref{tab:app_mastery_level} and \ref{tab:app_pervasiveness}) confirm this significant transfer persists even without the pervasiveness task and at $100\%$ initial accuracy. From an optimization perspective, post-training transitions the LLM from infinite to finite objectives, naturally shifting the optimal solution space. Ultimately, this dynamic nature confirms that static mechanistic conclusions (circuits) derived from current parameters cannot reliably represent future neural contributions.

To address \textbf{RQ1-b}, we track the evolution of attention ($W_q, W_k, W_v, W_o$; denoted \textit{Attn}) and MLP ($W_{\text{up}}, W_{\text{down}}$; denoted \textit{MLP}) components. As shown in Figure~\ref{figfreeevomistral}, \textit{Attn} components consistently exhibit significantly higher $CD$ than \textit{MLP} components throughout SFT. Prior mechanistic studies~\citep{meng2022locating, geva2021transformer} associate Attention with inter-token ``skills'' (e.g., induction heads dedicated to processing ``A B \dots A'' patterns, or syntax heads handling structures like ``The + Noun'') and MLPs with semantic ``knowledge'' (e.g., factual concepts). Combining these insights reveals a fundamental evolutionary pattern: ``skill''-centric \textit{Attn} circuits are highly volatile and prone to migration during updates, whereas ``knowledge''-centric \textit{MLP} circuits remain largely inert. This dichotomy is further corroborated by $CS$ trajectories, where \textit{MLP} stability substantially exceeds \textit{Attn} stability.
\begin{wraptable}{r}{0.7\linewidth}
\centering
\caption{Summary of key factors on free evolution}
\label{tab:ablation_summary}
\resizebox{\linewidth}{!}{
\begin{tabular}{lcccc}
    \toprule
    \textbf{Factor} & \textbf{$CD$ of Attn} & \textbf{$CD$ of MLP}& \textbf{$CS$ of Attn}&\textbf{$CS$ of MLP} \\
    \midrule
    ``Skill'' tasks &$\uparrow$ & - & $\downarrow$&- \\
    ``Knowledge'' tasks & -& $\downarrow$ &- &$\uparrow$ \\
   Pervasiveness $\uparrow$ &$\uparrow$ & $\uparrow$ &- &- \\
   Dataset Size $\uparrow$ &- & - &$\uparrow$ &$\uparrow$ \\
   Conflicting$\uparrow$&$\uparrow$ &$\uparrow$  & $\downarrow$&$\downarrow$ \\
   Mastered tasks&$\downarrow$ & $\downarrow$ &$\uparrow$ &$\uparrow$ \\
   Unmastered tasks&$\uparrow$ &$\uparrow$  & $\downarrow$&$\downarrow$\\
    \bottomrule
  \end{tabular}}
\end{wraptable}
This phenomenon is logically sound, given that MLPs contain vastly more parameters than Attention layers, making their stored knowledge more deeply entrenched and harder to migrate. We further validate the differential evolutionary impacts of ``skill'' versus ``knowledge'' tasks in the subsequent analysis of \textbf{RQ1-c}.

Finally, to address \textbf{RQ1-c}, we designed a series of ablation experiments to observe the impact of varying factors on Circuit Distance ($CD$) and Circuit Stability ($CS$). Ultimately, we identified five key factors that significantly influence circuit evolution:

\begin{itemize}
    \item \textbf{Task Type:} Skill-centric tasks (Attention-dominated) are highly susceptible to component migration, whereas knowledge-centric tasks (MLP-dominated) favor internal information updates (reflected in $CS$).
    \item \textbf{Degree of Pervasiveness:} Co-optimizing with increasingly pervasive tasks broadens component participation, triggering more extensive migration.
    \item \textbf{Dataset Size:} Larger SFT datasets intensify internal information refinement, enhancing knowledge consolidation and overall circuit robustness.
    \item \textbf{Conflict Proportion:} Higher ratios of conflicting components sharply increase migration and hinder effective information updates, diminishing circuit robustness.
    \item \textbf{Initial Mastery:} High initial task proficiency minimizes component migration and accelerates the efficiency of internal information updates.
\end{itemize}

Table~\ref{tab:ablation_summary} summarizes these factor-metric correlations, with comprehensive ablation data detailed in Appendix~\ref{suppfreeevo}.

\subsection{Localization Evolution}
\label{subsec:localization_evolution}

In this section, we transition to evaluating the dynamics of \textbf{Localization Evolution}. Maintaining the identical experimental setup as in Section~\ref{subsec:free_evolution}, we compare three distinct localization strategies during SFT:
\begin{itemize}
    \item \textbf{Free (Baseline):} Full-parameter SFT without freezing any parameters.
    \item \textbf{Mech (Mechanistic Localization):} Localizing critical components based on the initial parameter state ($\theta$) via mechanistic interpretability, while freezing all non-critical components.
    \item \textbf{Random:} Randomly localizing and updating the exact same number of components as identified in the \textit{Mech} strategy (same ratio between Attn and MLP), while freezing the remainder.
\end{itemize}

Through the comparative analysis of these three strategies, we aim to systematically address \textbf{RQ2} (\textit{Does post-training with localization truly improve performance on the target task and mitigate conflicts with pervasiveness tasks?}). To provide a granular analysis, we decompose this overarching inquiry into two specific sub-questions:
\begin{itemize}
    \item \textbf{RQ2-a:} Does static localization genuinely provide guidance for dynamic parameter updates?
    \item \textbf{RQ2-b:} Why do existing localization methods ostensibly yield significant improvements in task performance?
\end{itemize}

\subsubsection{RQ2-a}
To address \textbf{RQ2-a}, we compared three localization strategies (\textbf{Free}, \textbf{Mech}, and \textbf{Random}) on Mistral-7B. Consistent with previous settings, we paired five target tasks (Arithmetic, Bool, Gender, SST2, Winogrande) with the OpenBookQA pervasiveness task. For the \textit{Mech} strategy, we isolated $800$ critical components (the predetermined circuit scale). To ensure a rigorously controlled comparison, the \textit{Random} strategy correspondingly updated exactly $800$ randomly selected components.

\begin{figure*}
    \centering
    \subfigure[Target Accuracy]{
    \includegraphics[width=0.31\linewidth]{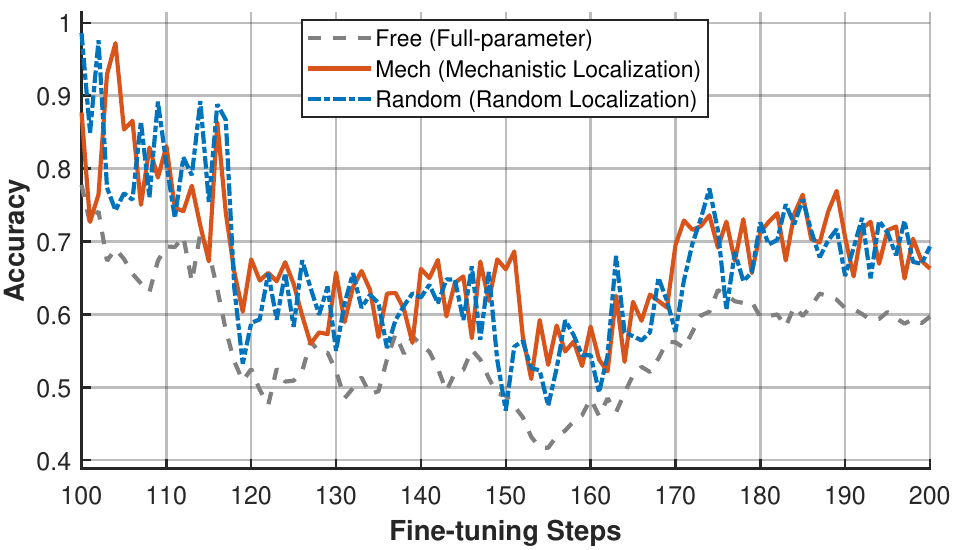}}
    \subfigure[Pervasiveness Accuracy]{
    \includegraphics[width=0.31\linewidth]{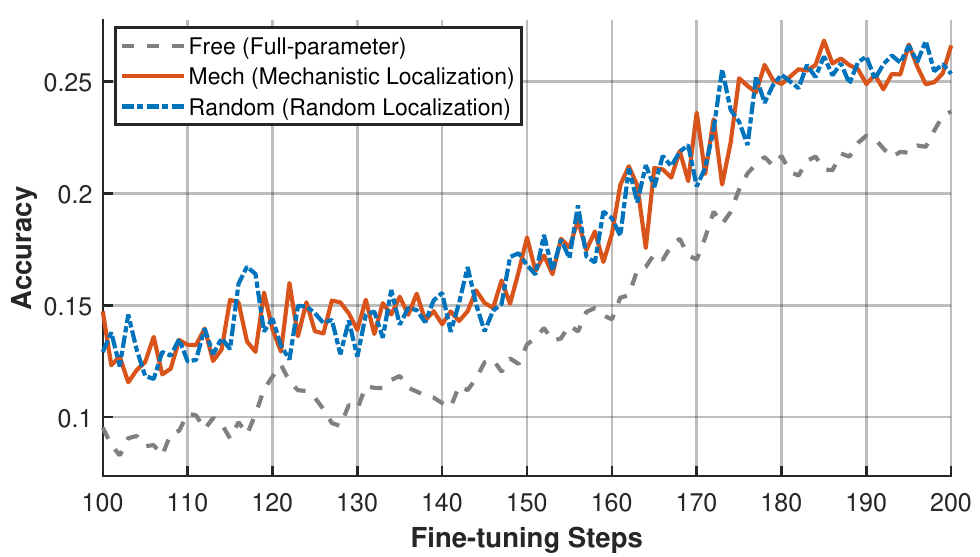}}
    \subfigure[Circuit Conflict]{
    \includegraphics[width=0.31\linewidth]{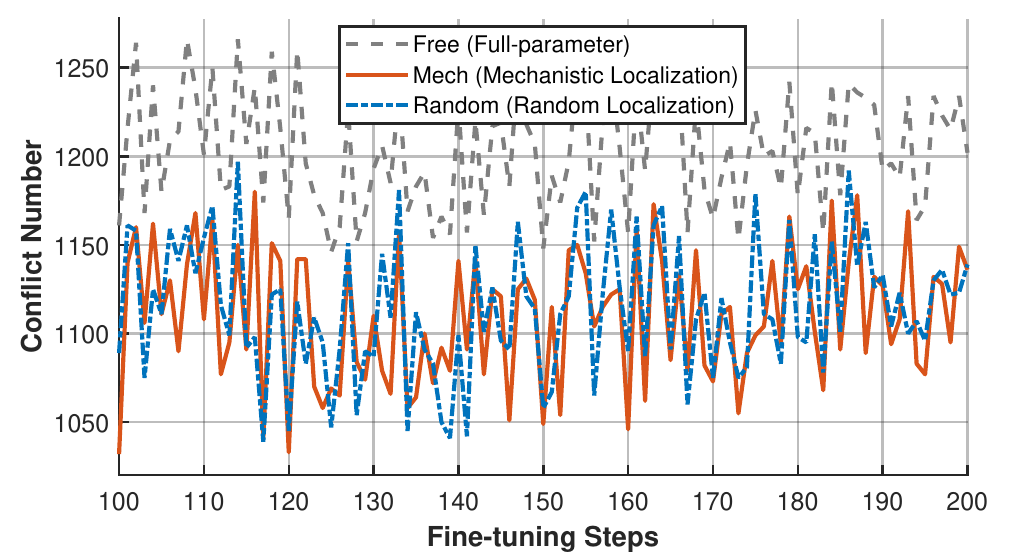}}
    \caption{Target Task Accuracy, Pervasiveness Task Accuracy, and Circuit Conflict of Arithmetic Task with localization. }
    \label{fig:rq2_arithmetic}
\end{figure*}

Figure~\ref{fig:rq2_arithmetic} illustrates the results for the Arithmetic task across three metrics: Target Task Accuracy, Pervasiveness Task Accuracy, and Circuit Conflict (The results of the other 4 tasks are shown in Appendix~\ref{moretaskinLocal}.). A compelling observation emerges: First, the Target Task Accuracy fails to surpass its pre-SFT (Supervised Fine-Tuning) baseline, a phenomenon that starkly contrasts with typical outcomes observed in single-objective optimization. Coupled with the evidence of circuit conflicts presented in Figure~\ref{fig:rq2_arithmetic} (c), this substantiates that conflicts arising from multi-task optimization inevitably degrade individual task performance. Furthermore, a compelling observation emerges: while the \textit{Mech} strategy significantly outperforms the \textit{Free} baseline in terms of accuracy and successfully maintains a lower conflict level, it exhibits no substantial advantage over the \textit{Random} strategy. This outcome reveals that while localization \textit{in general} does benefit task performance, this improvement ostensibly does not stem from the mechanistic interpretability guidance of the circuit.

The $CD$ and $CS$ metrics provided in Appendix~\ref{subsec:app_f2_cd_cs} further corroborate this: compared to free evolution, localization induces greater Circuit Distance and variations in Circuit Stability. In other words, localization paradoxically renders the circuit's evolution more uncontrollable and unstable, and the improvement in task performance may stem from the reduction in parameters, which makes task optimization simpler. This validates our preceding hypothesis: Mechanistic Localization based on the current parameter state identifies only a \textit{fraction} of the components that genuinely contribute to future parameter updates. Because the \textbf{unidentified critical components} are prematurely \textbf{frozen}, they impede normal evolutionary dynamics, rendering circuit evolution substantially more difficult. Consequently, from a macroscopic perspective, its performance is virtually indistinguishable from ``random localization.''Furthermore, Appendix~\ref{subsec:app_f3_circuit_scale} provides additional validation for this hypothesis. Expanding the circuit scale---thereby encompassing more ``\textbf{unidentified critical components}''---allows Mechanistic Localization to surpass Random Localization. 
\subsubsection{RQ2-b}\label{secrq2b}
However, prevalent downstream applications (e.g., LLM unlearning, Knowledge Editing) report Mechanistic Localization vastly outperforming Random Localization, starkly contradicting \textbf{RQ2-a}. We hypothesize this discrepancy arises because these applications predominantly target MLP-governed, knowledge-centric tasks. As established in \textbf{RQ1-b} (Section~\ref{subsec:free_evolution}), MLP circuits inherently resist component migration. Consequently, the divergence between current and updated circuits is significantly diminished. This minimal evolutionary drift creates an ``illusion of effectiveness''---the false premise that the ``current circuit'' reliably guides the ``future circuit.''

To empirically validate this hypothesis, we conducted a comparative analysis within the LLM unlearning paradigm, contrasting mainstream unlearning on the WMDP-Bio dataset against unlearning the Induction task. The WMDP-Bio dataset comprises biological knowledge and related factual information; hence, unlearning WMDP-Bio is a quintessential knowledge-centric task dominated by MLP circuits. Conversely, the Induction task relies on a skill-centric circuit predominantly driven by Attention mechanisms, making its unlearning an Attention-dominated process. 

We evaluated these two target tasks with OpenBookQA acting as the pervasiveness task. The comprehensive metrics tracked include ``FE'' (Forget efficacy) is measured as 1-accuracy on unlearning task and ``RU'' (Retain utility) is measured as accuracy of pervasiveness task, Circuit Distance of Attention ($CD_{Attn}$), Circuit Distance of MLP ($CD_{MLP}$), Circuit Stability of Attention ($CS_{Attn}$), Circuit Stability of MLP ($CS_{MLP}$), and Circuit Conflict ($CC$).

\begin{table}[ht]
  \caption{Performance of WMDP-Bio and Induction tasks on LLM unlearning}
  \label{tab:rq2b_unlearning_comparison}
  \centering
  \resizebox{0.9\textwidth}{!}{
  \begin{tabular}{lllllllll}
    \toprule
  Datasets & Localization & FE&RU&$CD_{Attn}$&$CD_{MLP}$&$CS_{Attn}$&$CS_{MLP}$&$CC$\\
    \midrule
     \multirow{3}{*}{WMDP-Bio}&Free&0.685&0.467&0.8412&0.5749&0.7819&0.9557&1055\\
                            &Random&0.716&0.441&1.1243&0.9557&0.5219&0.8396&1242\\
                              &Mech&0.736&0.457&1.0519&0.7391&0.6467&0.9082&1091\\
    \hline
    \multirow{3}{*}{Induction}&Free&0.873&0.324&1.2479&0.8949&0.7544&0.8846&1233\\
                             &Random&0.952&0.359&1.5866&1.0547&0.8137&0.8573&1153\\
                               &Mech&0.955&0.367&1.5593&1.1134&0.7968&0.8682&1129\\
    \bottomrule
  \end{tabular}}
\end{table}

Table~\ref{tab:rq2b_unlearning_comparison} corroborates our hypothesis: because the circuits of knowledge-centric tasks are largely composed of MLPs and their associated edges, the immense parameter volume and dense knowledge superposition within MLPs render them highly resistant to neural migration. In stark contrast, skill-centric tasks governed by Attention mechanisms are much more susceptible to interference due to their smaller parameter footprint, leading to a substantially higher proportion of component migration. In Appendix~\ref{appendix:single_objective}, we provide supplementary sampling results during the optimization process when these two tasks are treated as independent optimization objectives, further substantiating this conclusion. 

Therefore, from the perspective of circuit-guided parameter dynamics, the circuits of Attention-dominated skill tasks undergo drastic transformations during parameter updates. Mechanistic conclusions derived from these circuits suffer from severe temporal latency, rendering them fundamentally incapable of providing practical, meaningful contributions to future parameter updates.

\section{Discussion: The Path to Predictive Localization}
\label{sec:discussion}

Section~\ref{sec:experiments} demonstrates that task mechanisms inherently undergo ``free evolution.'' Due to these dynamic shifts, static interpretability conclusions suffer from severe temporal latency, fundamentally failing to guide future parameter updates. To make Mechanistic Localization practical, we must endow it with ``foresight''---the ability to predict future mechanisms.

\begin{figure*}
    \centering
    \subfigure[Task Accuracy]{
    \includegraphics[width=0.47\linewidth]{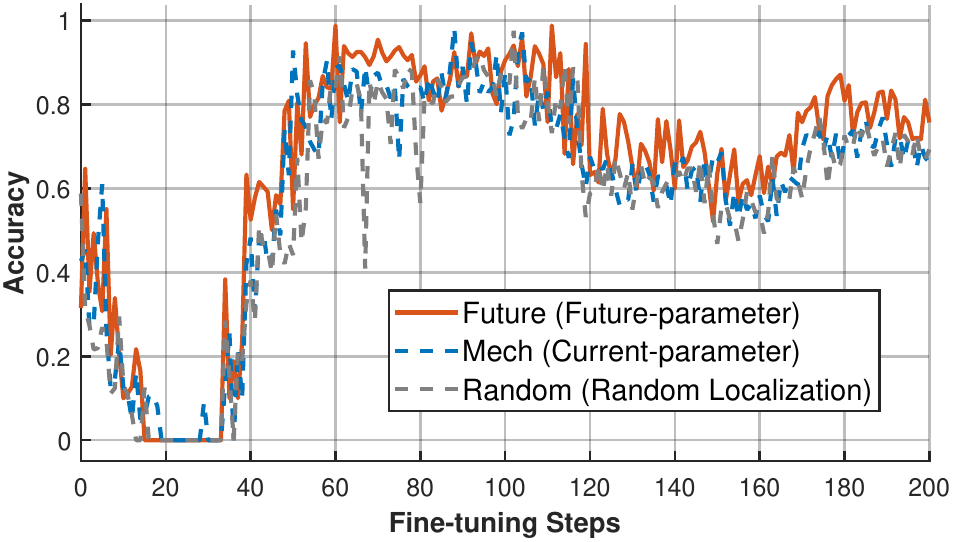}}
    \subfigure[Pervasiveness Accuracy]{
    \includegraphics[width=0.47\linewidth]{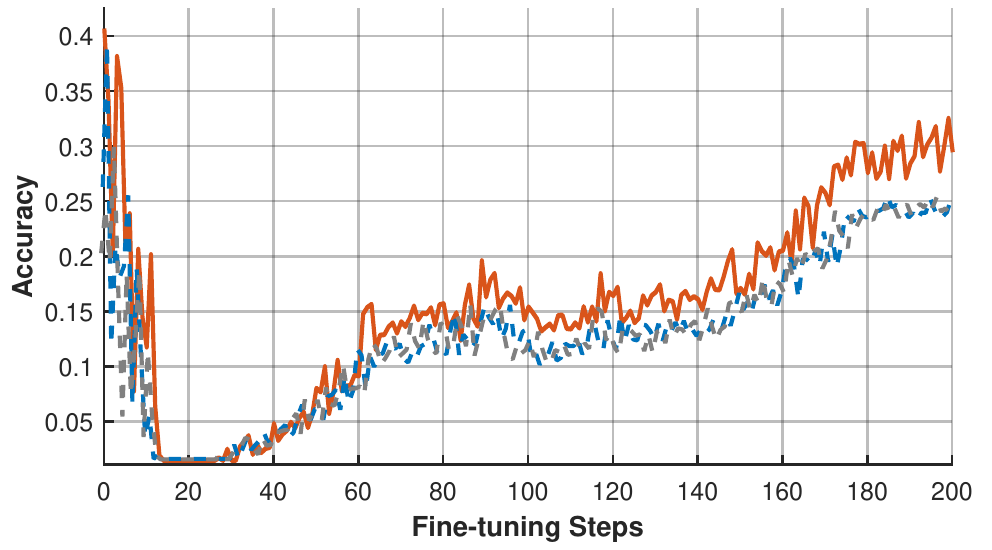}}
    \caption{Line plots of Future-Localization.}
    \label{fig:future_mechanistic}
\end{figure*}

Ideally, successfully forecasting the post-update circuit from the current parameter state would enable flawless critical component localization. Accordingly, we conducted a preliminary exploration using the circuit obtained \textit{after} free evolution SFT as a surrogate for the current circuit, introducing a paradigm we term \textit{Future Mechanistic} localization. Figure~\ref{fig:future_mechanistic} shows that this approach yields substantial advantages in both Target (T-Acc) and Pervasiveness (P-Acc) Accuracy on the Arithmetic task (additional metrics in Appendix~\ref{suppmorefuture}), powerfully corroborating our hypothesis and charting a clear trajectory for future research.

Furthermore, Appendix~\ref{suppmorefuture} presents a comprehensive analysis of contemporary Mechanistic Localization methodologies. Moving beyond strictly circuit-centric approaches, we evaluate two other predominant paradigms: gradient-based methods and intervention-based methods. Our empirical results reveal that critical components localized via gradient-based techniques exhibit markedly superior performance in terms of Circuit Distance ($CD$). This suggests a compelling insight: leveraging current gradient signals to approximate or extrapolate the mechanistic structures of future parameter states represents a highly promising frontier for the evolution of predictive Mechanistic Localization.

\section{Conclusion}
\label{sec:conclusion}

In this paper, we explore whether interpretability conclusions derived from current parameter states offer predictive guidance for future parameter updates. To investigate this, we constructed a ``Locate-then-Update'' pipeline using circuit discovery for Mechanistic Localization. By tracking circuits across SFT steps via three novel metrics---Circuit Distance ($CD$), Circuit Stability ($CS$), and Circuit Conflict ($CC$)---we systematically evaluated the evolutionary dynamics of task mechanisms. Extensive experiments substantiate three core conclusions:

\begin{itemize}
    \item \textbf{Divergent Free Evolution:} Without localization, circuits naturally undergo ``free evolution.'' Attention-driven skill tasks experience drastic structural shifts, whereas MLP-reliant knowledge tasks evolve much more gradually.
    \item \textbf{Temporal Lag of Static Localization:} Due to this inherent free evolution, utilizing current parameter states as a reference suffers from severe temporal latency. Consequently, static Mechanistic Localization fails to reliably guide future dynamic updates.
    \item \textbf{The Illusion of Effectiveness:} The perceived success of existing localization methods stems from their evaluation on MLP-dominated tasks. The inherently slower evolution of these tasks coincidentally masks the latency of static circuits, creating a false sense of efficacy.
\end{itemize}
Finally, we discuss potential research directions for achieving effective Mechanistic Localization and outline the limitations of this work in Appendix~\ref{supplimitation}.

\newpage
\bibliographystyle{unsrt}
\bibliography{reference}

%%%%%%%%%%%%%%%%%%%%%%%%%%%%%%%%%%%%%%%%%%%%%%%%%%%%%%%%%%%%

\appendix
\section{Details of Logical Circuit Framework}~\label{supplogicalcircuit}
At first, we systematically introduce three fundamental circuit logic types: the \textbf{AND} gate, \textbf{OR} gate, and \textbf{ADDER} gate~\citep{chen2025rethinking}.  

\begin{definition}
We assume a common paradigm in which a receiver node $B$, which is connected by more than 1 sender node $A_1, A_2, ...$. For any edge $A_i \rightarrow B$, we use binary values ‘0’ and ‘1’ to represent the activation state of a node. Specifically, $A_i=0$ indicates that node $A_i$ is removed, ablated, or deactivated, whereas $A_i=1$ indicates that node $A_i$ is retained and active. When the sender nodes are ablated, the effect of node $B$ on the output exhibits three distinct patterns, which are as follows: 

\textbf{AND}: All sender nodes satisfy an AND logical relationship with the receiver node, i.e., $B = A_1 \land A_2 \land \dots$. In this case, node $B$ exerts a significant effect on the output only if all of its sender nodes are retained. If even a single sender node is ablated, the effect of $B$ on the output is nearly eliminated. 

\textbf{OR} gate: All sender nodes satisfy an OR logical relationship with the receiver node, i.e., $B = A_1 \lor A_2 \lor \dots$.  In this case, node $B$ always exerts a significant effect on the output if one or more of its sender nodes are retained. Only if all sender nodes are ablated, the effect of $B$ on the output is nearly eliminated. 

\textbf{ADDER} gate: all sender nodes satisfy an ADDER logical relationship with the receiver node, i.e., $B = A_1 + A_2 + \dots$. In this case, node $B$ exhibits its maximal effect on the output only when all of its sender nodes are retained. If any single sender node is ablated, the effect of $B$ on the output is substantially diminished; when all sender nodes are ablated, $B$'s effect on the output is reduced to zero. Accordingly, we define the state of $B$ as taking values 0,1,2,., where the total number of distinct states equals the number of sender nodes. 
\label{defnitionlogicgate}
\end{definition}

Theoretical analyses support the view that noising-based intervention is capable of recovering a complete AND gate but fails to recover a complete OR gate, whereas denoising-based intervention demonstrates the opposite pattern~\citep{heimersheim2024use}. This asymmetry is straightforward to interpret. The noising-based intervention procedure corresponds to the transition from a clean activation state ($ \text{state}=1 $) to a corrupted activation state ($ \text{state}=0 $). Since all gates can be regarded as being initialized with activation states equal to $1$, any transition to $ \text{state}=0 $ induces a significant change in the effect of AND and ADDER gates on the output. Consequently, noising-based intervention can reliably identify AND and ADDER gates.

The \textbf{denoising-based intervention} first performs the corrupted run in the computational graph, and then replaces the corrupted activations with the clean activations. Those activations that lead to significant changes in the output ($\tilde{y}$) consist of the circuits. denoising-based intervention thus has the following objective: 
\begin{equation}
    \arg \min_{\mathcal{C}}\mathbb{E}_{(x,\tilde{x})\in \mathcal{T}}[D(p_{\mathcal{G}}(\tilde{y}|\tilde{x})||p_{\mathcal{C}}(\tilde{y}|\tilde{x},x))], ~~s.t.~1-|\mathcal{C}|/|\mathcal{G}|\geq s
    \label{eqtdenosing}
\end{equation}
Conversely, the denoising-based intervention procedure corresponds to initialization with activation states equal to $0$. In this case, any transition to $ \text{state}=1 $ produces a significant change in the effect of OR and ADDER gates on the output.

Therefore, we denote the circuit constructed under the noising-based intervention strategy as $\mathcal{C}_{\text{Ns}}$, and the one constructed under the denoising-based intervention strategy as $\mathcal{C}_{\text{Dn}}$. Based on the above set-theoretic relationships between $\mathcal{C}_{\text{Ns}}$ and $\mathcal{C}_{\text{Dn}}$, we extract subsets of edges corresponding to AND, OR, and ADDER gates as follows:
\begin{itemize}
    \item AND gate ($\mathcal{C}_{\text{AND}}$): edges that are present in $\mathcal{C}_{\text{Ns}}$ but absent from $\mathcal{C}_{\text{Dn}}$.
    \item OR gate ($\mathcal{C}_{\text{OR}}$): edges that are present in $\mathcal{C}_{\text{Dn}}$ but absent from $\mathcal{C}_{\text{Ns}}$.
    \item ADDER gate ($\mathcal{C}_{\text{ADDER}}$): edges that are shared between $\mathcal{C}_{\text{Ns}}$ and $\mathcal{C}_{\text{Dn}}$.
\end{itemize} 

Therefore, we propose a combined \textbf{Ns+Dn} approach to recover logically complete gates. This method is compatible with a wide range of circuit discovery algorithms, introduces minimal additional computational overhead, and enables clear and effective separation of the three types of logic gates. Ns+Dn has the following objective: 
\begin{equation}
    \arg \min_{\mathcal{C}}\mathbb{E}_{(x,\tilde{x})\in \mathcal{T}}[D(p_{\mathcal{G}}(y|x)||p_{\mathcal{C}}(y|x,\tilde{x}))+D(p_{\mathcal{G}}(\tilde{y}|\tilde{x})||p_{\mathcal{C}}(\tilde{y}|\tilde{x},x))], ~~s.t.~1-|\mathcal{C}|/|\mathcal{G}|\geq s
    \label{eqtcomplete}
\end{equation}
 In the following sections, we provide a detailed exposition of the original design of each method under the Ns. strategy, the corresponding formulation under the Dn. strategy, and the final approach that integrates both—Ns.+Dn.—for recovering logically complete gates.

 \subsection{Greedy Search Example: ACDC}
The ACDC method identifies important edges by iteratively removing each edge and observing the effect of this intervention on the model output. Edges whose removal causes an effect greater than a predefined threshold $\tau$ are retained, while those with an effect smaller than $\tau$ are pruned. The original algorithm (Ns. strategy), is outlined as follows: 
\begin{algorithm}
\caption{The ACDC algorithm in Ns.}\label{acdc_algorithmns}
\DontPrintSemicolon % Removes semicolons
\KwData{Computational graph $\mathcal{G}$, dataset $(x_i)_{i=1}^n$, corrupted datapoints $(x_i')_{i=1}^n$ and threshold $\tau > 0$.}
\KwResult{Subgraph $\mathcal{H} \subseteq \mathcal{G}$.}
$\mathcal{H} \gets \mathcal{G}$ \tcp*[r]{Initialize H to the full computational graph}
$\mathcal{H} \gets \mathcal{H}.{reverse\_topological\_sort()}$\tcp*[r]{Sort H so output first}\label{line:sort}
\For{$v \in \mathcal{H}$}{
  \For{$w$ parent of $v$}{ \label{line:parents}
    $\mathcal{H}_\mathrm{new} \gets \mathcal{H} \setminus \{w \rightarrow v\}$\tcp*[r]{Temporarily remove candidate edge}
    \If{$D_{KL}(\mathcal{G} || \mathcal{H}_\mathrm{new}) - D_{KL}(\mathcal{G} || \mathcal{H}) < \tau$ \label{line:kl}}
    % \If{$F(H_\mathrm{new}) - F(H) < \tau$ \label{line:kl}}
    {
      $\mathcal{H} \gets \mathcal{H}_\mathrm{new}$\tcp*[r]{Edge is unimportant, remove permanently}
    }
  }
}
\Return{$\mathcal{H}$}
\end{algorithm}

In the \textbf{Ns.} strategy, $\mathcal{G}$ denotes the \textbf{clean run}, and $\mathcal{H} \setminus \{w \rightarrow v\}$ represents the replacement of the clean activation on the edge $w \rightarrow v$ with its corrupted activation. In contrast, under the \textbf{Dn.} strategy, $\mathcal{G}$ refers to the \textbf{corrupted run}, and $\mathcal{H} \setminus \{w \rightarrow v\}$ indicates the substitution of the corrupted activation on edge $w \rightarrow v$ with the corresponding clean activation.

In the combined \textbf{Ns.+Dn.} approach, the effects from both strategies are jointly considered. Specifically, the original pruning condition  
$D_{KL}(\mathcal{G} \,\|\, \mathcal{H}_\mathrm{new}) - D_{KL}(\mathcal{G} \,\|\, \mathcal{H}) < \tau$
is replaced with the aggregated criterion:  
$D_{KL}(\mathcal{G}^{\text{clean}} \,\|\, \mathcal{H}_\mathrm{new}) - D_{KL}(\mathcal{G}^{\text{clean}} \,\|\, \mathcal{H}) + D_{KL}(\mathcal{G}^{\text{corrupted}} \,\|\, \mathcal{H}_\mathrm{new}) - D_{KL}(\mathcal{G}^{\text{corrupted}} \,\|\, \mathcal{H}) < \tau.$

\subsection{Linear Estimation Example: EAP}
The EAP method approximates the effect of each edge using the first-order term of its Fourier expansion, enabling the estimation of all edge effects with a single forward pass. It is important to note that, during the computation of each edge’s effect, all other edges remain in their unpruned (active) state.

Specifically, Ns. has approximation: 
\begin{equation}
    L(x|do(\tilde{x}_{i}))-L(x)\approx(\tilde{x}_{i}-x_{i})^{T}\frac{\partial}{\partial x_{i}}L(x)
\end{equation}

and Dn. has approximation: 
\begin{equation}
    L(\tilde{x}|do(x_{i}))-L(\tilde{x})\approx(\tilde{x}_{i}-x_{i})^{T}\frac{\partial}{\partial \tilde{x}_{i}}L(\tilde{x})
\end{equation}
Therefore, the approximation for Ns.+Dn. is $(\tilde{x}_{i}-x_{i})^{T}\frac{\partial}{\partial x_{i}}L(x)+(\tilde{x}_{i}-x_{i})^{T}\frac{\partial}{\partial \tilde{x}_{i}}L(\tilde{x})$.
\subsection{Differentiable Mask Example: EdgePruning}\label{suppedgepruning}
EdgePruning assigns a learnable mask to each node or edge, where the mask is reparameterized using the hard concrete distribution. In the Ns. setting, the optimization objective corresponds to Equation~\ref{eqtnosing}. Consequently, the objectives for the Dn. and Ns.+Dn. settings are given by Equation~\ref{eqtdenosing} and Equation~\ref{eqtcomplete}, respectively.

In the Ns.+Dn. setting, directly optimizing both objectives jointly can lead to gradient interference and convergence to Pareto-optimal solutions, rather than a unified optimum. To address this, we independently compute the final mask values for Ns. and Dn. using Equations~\ref{eqtnosing} and~\ref{eqtdenosing}, and then obtain the mask for Ns.+Dn. by averaging the two. 
 
Finally, we simplify the ADDER gate in the forget circuit to an OR gate, and the ADDER gate in the retain circuit to an AND gate. 

\section{Experiment Details}\label{suppexdetail}
The learning rate is grid-searched at $1 \times 10^{-5}$ for each dataset. The parameter $\lambda=1$, and we adopted AdamW~\citep{loshchilov2017decoupled} as the optimizer.  All experiments were conducted on 16 NVIDIA RTX A100 GPUs.

We select a series of specific tasks as target and pervasiveness set: OpenBookQA~\citep{OpenBookQA2018}, Gender~\citep{mathwin2023identifying}, RTE~\citep{dagan2022recognizing}, IOI (Indirect Object Identification~\citep{wanginterpretability}), Docstring~\citep{heimersheim2023circuit}, SST2~\citep{socher-etal-2013-recursive}, Winogrande~\citep{ai2:winogrande}, Reverse~\citep{lindner2023tracr}, Greater Than~\citep{hanna2023does}, FEVER~\citep{thorne-etal-2018-fever}, zsRE~\citep{levy-etal-2017-zero}, Induction~\citep{conmy2023towards}, Bool~\citep{suzgun2023challenging}, Arithmetic~\citep{brown2020language}, and SA ( syntactic agreement~\citep{yu2024functional}). We show the examples of each task in the Table~\ref{tabretainexample}.

\begin{table}[ht]
  \caption{An overview of the datasets of specific tasks.}
  \label{tabretainexample}
  \centering
  \resizebox{0.9\textwidth}{!}{
  \begin{tabular}{lll}
    \toprule
    \textbf{Task}     & \textbf{Example}     & \textbf{Label}\\
    \midrule
   Winogrande&John moved the couch from the garage to the backyard to create space. The $\_$ is small.& garage \\
     \hline
   SST-2&hide new secretions from the parental units&negative\\
   \hline
   \multirow{2}{*}{RTE}&No Weapons of Mass Destruction Found in Iraq Yet.&\multirow{2}{*}{not entailment}\\
   &Weapons of Mass Destruction Found in Iraq.&\\
    \hline
   Bool&(True AND True) OR False & True\\
    \hline
   Induction&Vernon Dursley and Petunia Durs&ley\\
    \hline
   IOI&When John and Mary went to the store, Mary gave a bottle of milk to&John\\
    \hline
   Gender&So Evan is a really great friend, isn't&he\\
    \hline
   \multirow{4}{*}{Docstring}&def f(self, files, obj, state, size, shape, option):&\multirow{4}{*}{shape}\\
   &:param state: performance analysis&\\
   &:param size: pattern design&\\
   &:param&\\
    \hline
   Great Than&The war lasted from 1517 to 15&18\\
    \hline
   SA&Many girls insulted&themselves\\
    \hline
   arithmetic&What is (2 - 8) - 4? Answer: &-10\\
    \hline
   Reverse&[0, 3, 2, 1]&[1, 2, 3, 0]\\
     \hline
   \multirow{6}{*}{openbookqa}&'question$\_$stem': 'The sun is responsible for',&\multirow{6}{*}{D}\\
   &'choices': $\{$'text': ['puppies learning new tricks',&\\
   &'children growing up and getting old',&\\
   &'flowers wilting in a vase',&\\
   &'plants sprouting, blooming and wilting'],&\\
   &'label': ['A', 'B', 'C', 'D']$\}$,&\\
   \hline
   FEVER&Kolhan is a village in the Palghar district of Maharashtra , &India\\
   \hline
   zsRE&Angela Merkel’s second husband is professor Joachim&Sauer\\
    \bottomrule
  \end{tabular}}
\end{table}

\section{Ablation Studies and Additional Results on Free Evolution}
\label{suppfreeevo}

% C.1
\subsection{Comprehensive Results of Free Evolution}
\label{suppfreeevolineplot}

\begin{figure*}
    \centering
    \subfigure[Circuit Distance on Mistral]{
    \includegraphics[width=0.46\linewidth]{MistralFreeevoCD.pdf}}
    \subfigure[Circuit Stability on Mistral]{
    \includegraphics[width=0.46\linewidth]{MistralFreeevoCS.pdf}}
    \subfigure[Performance on Mistral]{
    \includegraphics[width=0.46\linewidth]{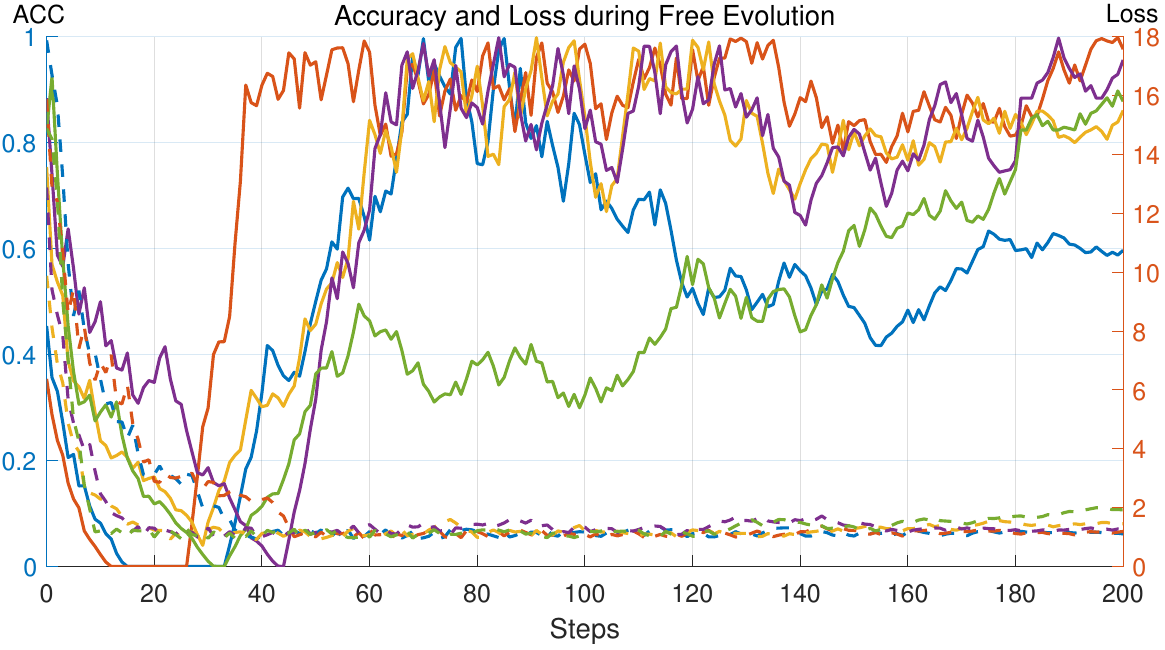}}
    \subfigure[Circuit Distance on LlaMA]{
    \includegraphics[width=0.46\linewidth]{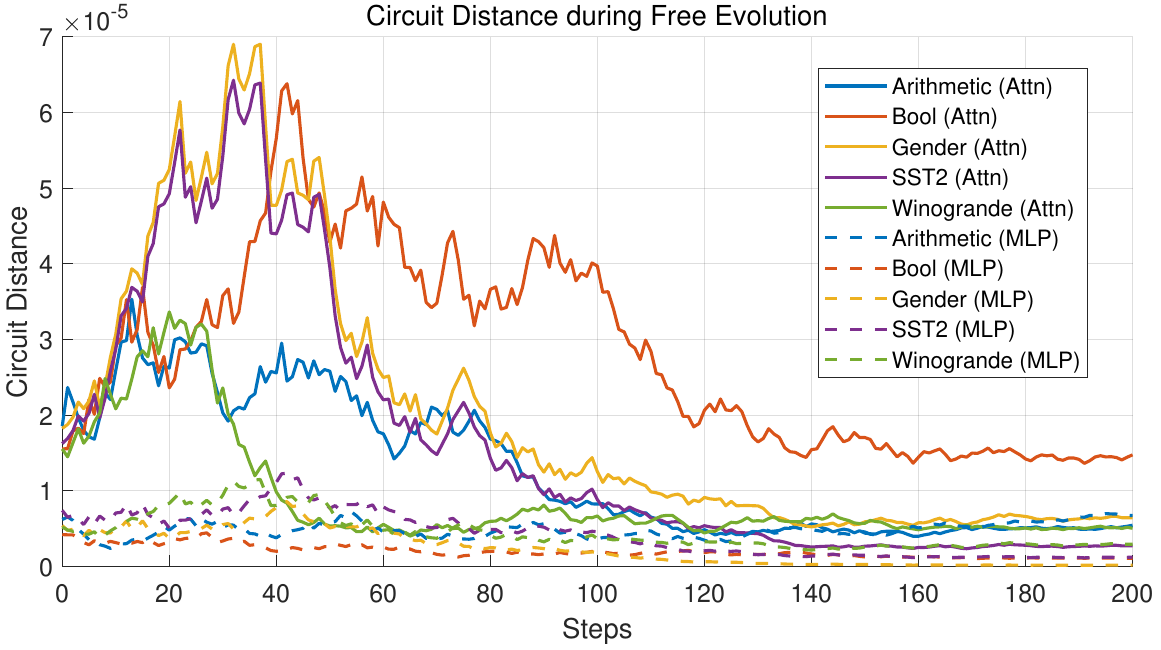}}
    \subfigure[Circuit Stability on LlaMA]{
    \includegraphics[width=0.46\linewidth]{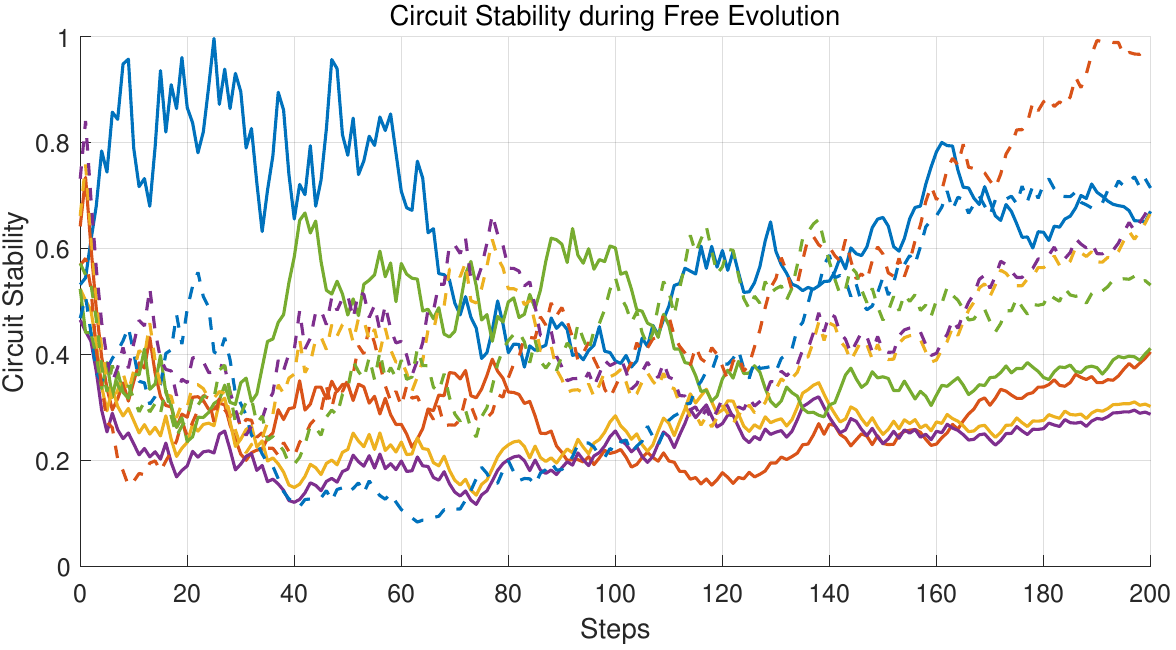}}
    \subfigure[ Performance on LlaMA]{
    \includegraphics[width=0.46\linewidth]{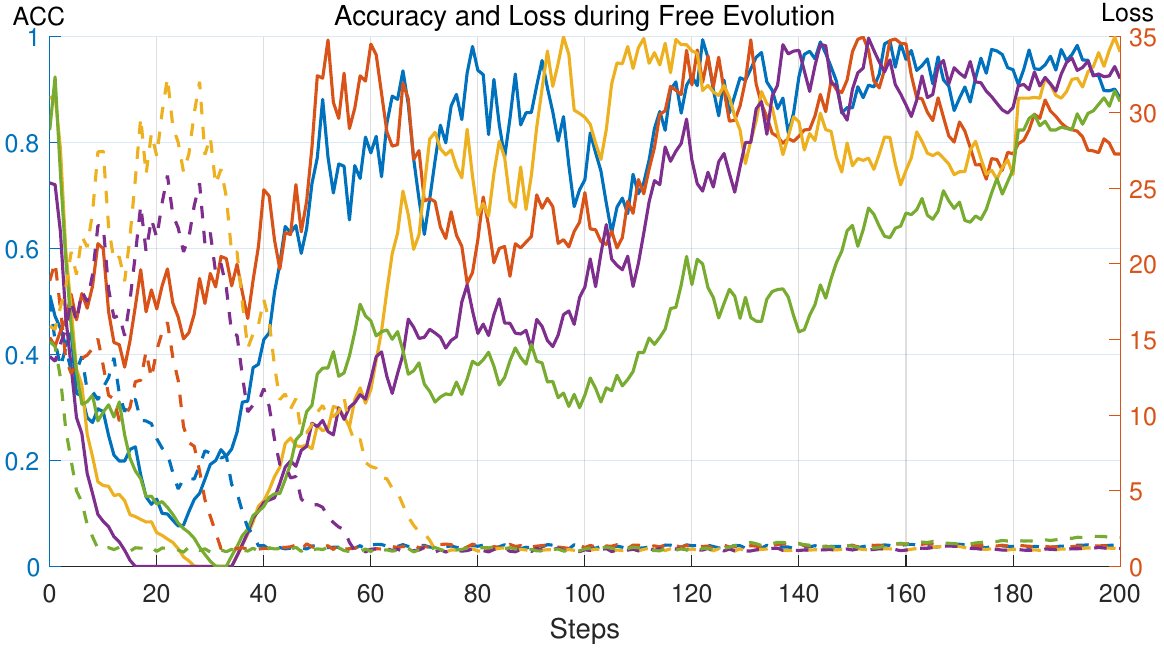}}
    \caption{line plots of different target tasks on the Mistral-7B model and LlaMA3-8B model in terms of Circuit Distance ($CD$),  Circuit Stability ($CS$), Task Accuracy ($ACC$) and loss ($Loss$).}
    \label{fig:app_free_evo_full}
\end{figure*}

Figure~\ref{fig:app_free_evo_full} presents the comprehensive evolutionary metrics for five target tasks (Arithmetic, Bool, Gender, SST2, and Winogrande) paired with OpenBookQA as the pervasiveness task, evaluated on both the Mistral-7B and LLaMA3-8B models. These tracked metrics encompass Circuit Distance ($CD$), Circuit Stability ($CS$), Task Accuracy ($ACC$), and Loss. Furthermore, for $CD$ and $CS$, we separately delineate the evolutionary dynamics of components within the attention and MLP modules, explicitly denoted as $Attn$ and $MLP$, respectively.

\subsection{Impact of Task Type on Free Evolution}
\label{subsec:app_task_type}

As observed in Figure~\ref{fig:app_free_evo_full}, components within the attention and MLP modules exhibit markedly distinct evolutionary patterns. Consequently, we hypothesize that for any given task, the extent of circuit variation during free evolution is heavily influenced by the relative functional contributions of its attention and MLP components. To validate this hypothesis, we designed the following ablation study. 

Prior research has established that tasks dominated by MLP components are predominantly associated with stored factual ``knowledge'' ~\citep{meng2022locating}(e.g., factual queries regarding nations, geography, prominent figures, and sports), whereas tasks dominated by attention components are generally tied to latent functional ``skills'' (e.g., induction heads~\citep{olsson2022context} and syntax heads~\citep{liu2026an}). Building upon this consensus, we select the \textit{Induction} and \textit{Reverse} tasks to represent attention-dominated (skill-centric) tasks, and the \textit{FEVER} and \textit{zsRE} factual datasets to represent MLP-dominated (knowledge-centric) tasks. 

Table~\ref{tab:app_task_type} details the performance across various metrics for these disparate task types on the Mistral-7B model, maintaining OpenBookQA as the pervasiveness task. The Attention and MLP columns represent the percentage of attention/MLP components in the circuit relative to the total number of components in the corresponding computational graph. The results substantiate our initial hypothesis: skill-centric tasks induce substantial migration among attention components, but cause negligible changes in MLP circuit distance. Conversely, knowledge-centric tasks trigger drastic internal parameter updates within MLPs (reflected in stability changes) but result in minimal positional migration of the circuit.

\begin{table}[ht]
  \caption{Performance of skill-centric tasks and knowledge-centric tasks}
  \label{tab:app_task_type}
  \centering
  \resizebox{1\textwidth}{!}{
  \begin{tabular}{llllllllll}
    \toprule
  Ablation & dataset & Attention($\%$) &MLP($\%$)&$\Delta$ ACC&Loss&$CD$ of Attn ($\times 10^{-5}$) &$CD$ of MLP ($\times 10^{-5}$) &$CS$ of Attn &$CS$ of MLP\\
    \midrule
     \multirow{2}{*}{Skill}&Induction&75.41&6.25&23.55	&1.4217	&1.6471	&0.8445	&0.6577	&0.8471\\
     &Reverse&64.33&9.375&33.6&	1.2471&	1.5421&	0.7419&	0.5944&	0.7619\\
     \multirow{2}{*}{Knowledge}&FEVER&33.71&40.625&15.47&	1.1945&	1.1474&	0.4571&	0.8847&	0.9492\\
     &zsRE&36.94&56.25&18.47&	1.2108&	1.2019&	0.4138&	0.8570&	0.9355\\
    \bottomrule
  \end{tabular}}
\end{table}

\subsection{Impact of Pervasiveness Degree on Free Evolution}
\label{subsec:app_pervasiveness}

In the context of LLMs, the optimization of any target task is inherently a multi-objective process accompanied by pervasiveness tasks. Therefore, we conducted an ablation study to investigate the impact of the degree of pervasiveness. Specifically, we selected various concrete tasks, each governing a distinct mechanism, and simulated increasing pervasiveness through their linear superposition (e.g., an ensemble of $10$ disparate tasks functioning jointly as the pervasiveness task intrinsically embodies a more ``universal'' constraint than a single isolated task). Accordingly, we chose Gender, RTE, IOI, Docstring, SST2, Winogrande, Reverse, Greater Than, FEVER, and zsRE as candidate components for the pervasiveness task. Designating Induction as the target task, we observed the circuit evolution on the Mistral-7B model when the total number of concurrent pervasiveness tasks was scaled to $0$, $1$, $2$, $5$, and $10$.

Table~\ref{tab:app_pervasiveness} demonstrates that as the degree of pervasiveness increases, the target task exhibits more extensive component migration, while the internally learned information remains largely unaffected. Intuitively, elevated pervasiveness implies a more comprehensive engagement of components across the network during parameter updates, equating to a more constrained and challenging optimization objective. Consequently, the original critical components of the target task typically require more substantial migration to navigate toward the new optimal solution space.
\begin{table}[ht]
  \caption{Performance of different pervasiveness}
  \label{tab:app_pervasiveness}
  \centering
  \resizebox{0.9\textwidth}{!}{
  \begin{tabular}{lllllll}
    \toprule
   task number&$\Delta$ ACC&Loss&$CD$ of Attn ($\times 10^{-5}$) &$CD$ of MLP ($\times 10^{-5}$) &$CS$ of Attn &$CS$ of MLP\\
    \midrule
     0&38.99&	0.9544&	1.4271&	0.6355&	0.6452&	0.6849\\
     1&33.51&	1.1146&	1.4533&	0.6411&	0.5391&	0.7145\\
     2&31.24&	1.1139&	1.4951&	0.6875&	0.5216&	0.7351\\
     5&25.17&	1.1547&	1.6742&	0.7422&	0.5479&	0.6557\\
     10&23.11&	1.3049&	1.8549&	0.8617&	0.5311&	0.6647\\
    \bottomrule
  \end{tabular}}
\end{table}

\subsection{Impact of Dataset Size on Free Evolution}
\label{subsec:app_dataset_size}

Similarly, we conducted an ablation study on the impact of dataset size on circuit evolution. We hypothesized that circuit migration might simply be an artifact of insufficient sample sizes, which could lead to uncertain or unstable sampled distributions. To isolate the effect of dataset size while simultaneously eliminating the confounding factor of sample imbalance, we designated the IOI task as the target task. The IOI dataset is a synthetic (procedurally generated) dataset, alleviating concerns that scaling up data samples might introduce real-world biases. Using the Mistral-7B model with OpenBookQA as the pervasiveness task, we evaluated circuit metrics across varying SFT dataset sizes: $500$; $2,000$; $5,000$; $10,000$; and $100,000$.

Table~\ref{tab:app_dataset_size} indicates that variations in dataset size do not induce significant changes in Circuit Distance ($CD$); that is, component migration during free evolution is largely independent of dataset size. However, dataset size profoundly impacts the updating of internal information within components. A richer set of training samples facilitates more effective internal information updating, thereby substantially enhancing the robustness of the resulting circuit.

\begin{table}[ht]
  \caption{Performance of different dataset size}
  \label{tab:app_dataset_size}
  \centering
  \resizebox{0.9\textwidth}{!}{
  \begin{tabular}{lllllll}
    \toprule
   sample number&$\Delta$ ACC&Loss&$CD$ of Attn ($\times 10^{-5}$) &$CD$ of MLP ($\times 10^{-5}$) &$CS$ of Attn &$CS$ of MLP\\
    \midrule
     500&28.45&	1.2541&	1.5749&	0.6645&	0.5517&	0.7541\\
     2000&31.41&	1.2037&	1.5422&	0.6217&	0.5916&	0.7594\\
     5000&32.04&	1.2316&	1.5594&	0.6559&	0.6471&	0.8051\\
     10000&32.57&	1.1842&	1.5207&	0.6347&	0.6689&	0.8067\\
     100000&36.14&	1.1554&	1.4971&	0.6467&	0.6953&	0.8137\\
    \bottomrule
  \end{tabular}}
\end{table}

\subsection{Impact of Conflict Proportion on Free Evolution}
\label{subsec:app_conflict_proportion}

Prior research has established that the quantity of conflicting components among multiple optimization tasks is a critical determinant of multi-task optimization efficacy~\citep{chen2025clue}. Accordingly, we constructed various skill combinations to serve as the pervasiveness task, aiming to observe how an escalating proportion of conflicting components impacts circuit evolution. In our experimental setup, we evaluated the Mistral-7B model using Gender as the target task. We formulated five distinct pervasiveness task combinations that yielded varying proportions of conflicting components relative to the target task, calculated following the methodology in~\citep{chen2025clue}. These combinations and their corresponding conflict proportions are: $5.86\%$ (SST2 + SA), $8.57\%$ (SST2 + SA + RTE), $10.17\%$ (SST2 + SA + RTE + IOI), $12.49\%$ (SST2 + SA + RTE + IOI + Winogrande), and $15.63\%$ (SST2 + SA + RTE + IOI + FEVER).

Table~\ref{tab:app_conflict_proportion} reveals that an increased proportion of conflicting components indeed precipitates significant component migration. As corroborated by previous studies, conflicting components are predominantly polysemantic, encapsulating essential information for multiple tasks simultaneously. When confronted with multi-task optimization, these components struggle to resolve optimal gradient descent directions. Consequently, the language model is compelled to disentangle these multiplexed semantics, forcing them to migrate into distinct, separate components to satisfy the diverse optimization objectives. Inevitably, this extensive migration necessitates a reorganization of internal neuronal semantics, which subsequently impedes the effective updating of internal information.

\begin{table}[ht]
  \caption{Performance of different conflict percentage}
  \label{tab:app_conflict_proportion}
  \centering
  \resizebox{0.9\textwidth}{!}{
  \begin{tabular}{lllllll}
    \toprule
   percentage($\%$)&$\Delta$ ACC&Loss&$CD$ of Attn ($\times 10^{-5}$) &$CD$ of MLP ($\times 10^{-5}$) &$CS$ of Attn &$CS$ of MLP\\
    \midrule
     5.86&29.63&	1.2346&	1.4377&	0.6147&	0.5846&	0.7841\\
     8.57&28.46&	1.3544&	1.4872&	0.6355&	0.5739&	0.7694\\
     10.17&26.17&	1.3927&	1.5239&	0.6741&	0.5517&	0.7614\\
     12.49&24.39&	1.4143&	1.5567&	0.6964&	0.5228&	0.7259\\
     15.63&22.15&	1.5467&	1.6741&	0.6857&	0.4677&	0.7118\\
    \bottomrule
  \end{tabular}}
\end{table}

% C.6
\subsection{Impact of Initial Mastery Level on Free Evolution}
\label{subsec:app_mastery_level}

Finally, we investigate the impact of the model's intrinsic mastery level of the target skill. This investigation is driven by a compelling hypothesis: if a model has already attained $100\%$ mastery of a task's mechanism (i.e., achieving $100\%$ accuracy), will its corresponding circuit still undergo evolution during subsequent parameter updates? 

To explore this, we selected a set of high-mastery tasks (Gender, SST2, Winogrande) and low-mastery tasks (Bool, Arithmetic, SA). High-mastery tasks are defined as those exhibiting a pre-SFT accuracy exceeding $70\%$, whereas low-mastery tasks fall below $40\%$. Furthermore, we curated a specialized subset comprising $500$ correctly answered samples extracted from the high-mastery datasets, ensuring a rigorous initial accuracy of $100\%$. Designating these datasets as target tasks and OpenBookQA as the pervasiveness task, we conducted ablation experiments on the Mistral-7B model.

Table~\ref{tab:app_mastery_level} demonstrates that tasks with higher initial mastery exhibit a diminished degree of component migration, suggesting that the model has already partially solidified the underlying mechanism. Crucially, however, even for skills mastered at $100\%$, the circuit still undergoes definitive migration and evolution during parameter updates. This finding compellingly reinforces the conclusion that a circuit is an inherently dynamic property; therefore, a static circuit derived from current parameters cannot reliably dictate or guide future parameter updates.

\begin{table}[ht]
  \caption{Performance of different mastery}
  \label{tab:app_mastery_level}
  \centering
  \resizebox{1\textwidth}{!}{
  \begin{tabular}{llllllll}
    \toprule
   ablation&dataset&$\Delta$ ACC&Loss&$CD$ of Attn ($\times 10^{-5}$) &$CD$ of MLP ($\times 10^{-5}$) &$CS$ of Attn &$CS$ of MLP\\
    \midrule
     \multirow{3}{*}{Mastered}&Gender&3.37&	1.2417&	0.8947&	0.9517&	0.6759&	0.8472\\
     &SST2&22.51&	1.1694&	1.1167&	0.6646&	0.8473&	0.9618\\
     &Winogrande&14.79&	1.2547&	0.9382&	0.5938&	0.7882&	0.8611\\
     \hline
     \multirow{3}{*}{100$\%$ Mastered}&Gender&-2.71&	1.1544&	0.8017&	0.8851&	0.8543&	0.8867\\
     &SST2&-6.88&	1.1392&	1.0544&	0.7419&	0.8144&	0.9546\\
     &Winogrande&-5.93&	1.1582&	1.0649&	0.8244&	0.7967&	0.9386\\
     \hline
     \multirow{3}{*}{Unmastered}&Bool&39.45&	1.3347&	1.3547&	1.1124&	0.4985&	0.6719\\
     &Arithmetic&32.57&	1.3629&	1.6691&	0.6749&	0.4467&	0.6637\\
     &SA&29.67&	1.2984&	1.8932&	0.9839&	0.5278&	0.6565\\
    \bottomrule
  \end{tabular}}
\end{table}

\section{More Experiments Results with Localization}

\subsection{More Target Tasks on Three Localization Strategies}\label{moretaskinLocal}

\begin{figure*}
    \centering
    \subfigure[T-ACC of Bool]{
    \includegraphics[width=0.31\linewidth]{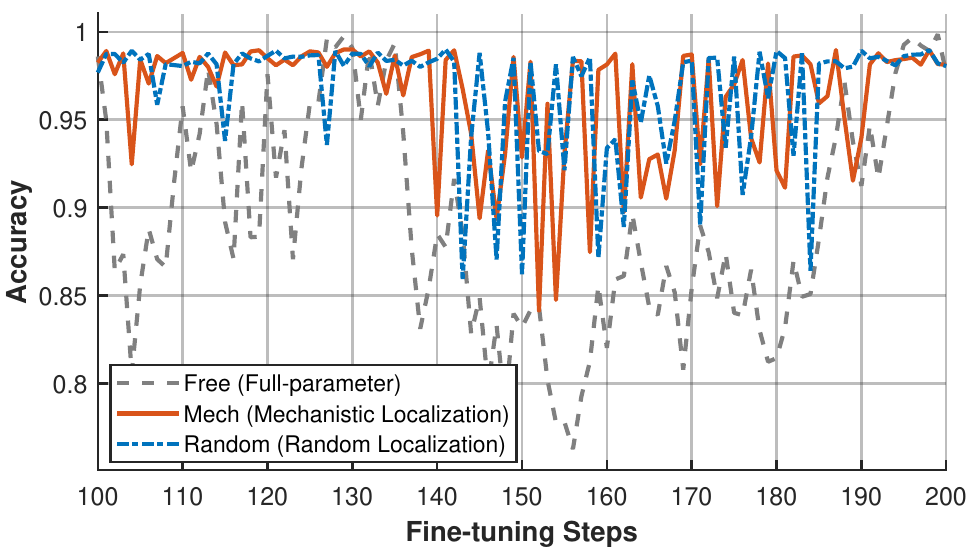}}
    \subfigure[P-Acc of Bool]{
    \includegraphics[width=0.31\linewidth]{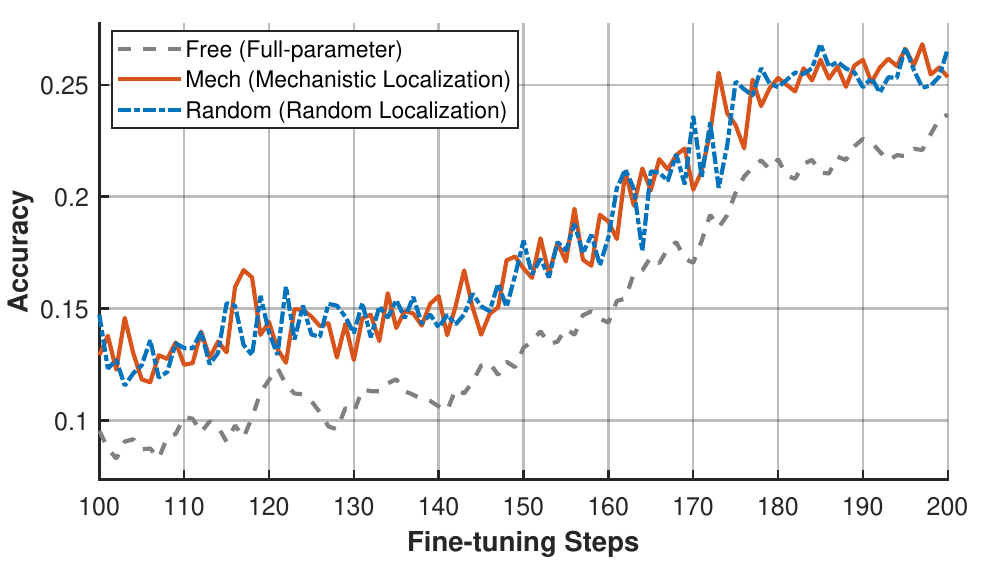}}
    \subfigure[CC of Bool]{
    \includegraphics[width=0.31\linewidth]{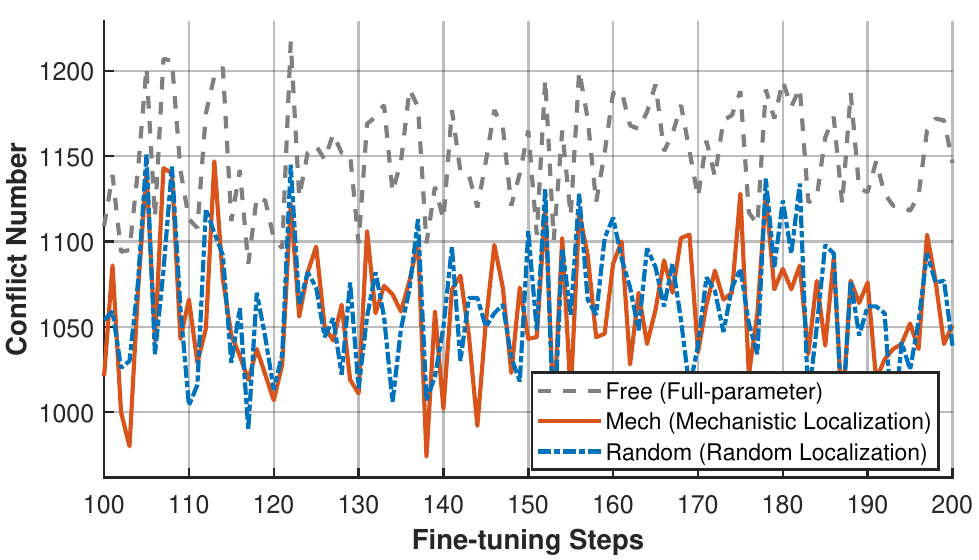}}

    \subfigure[T-ACC of Gender]{
    \includegraphics[width=0.31\linewidth]{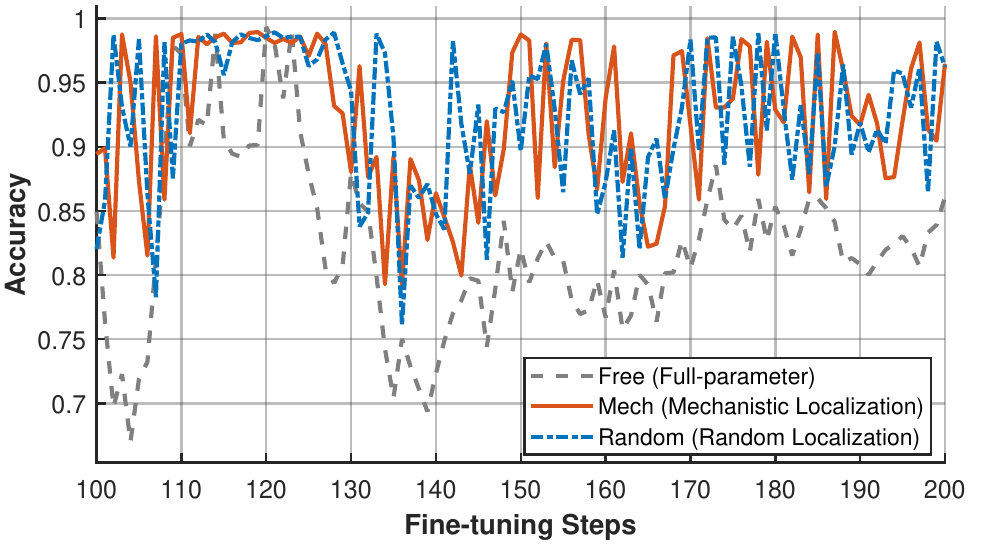}}
    \subfigure[P-Acc of Gender]{
    \includegraphics[width=0.31\linewidth]{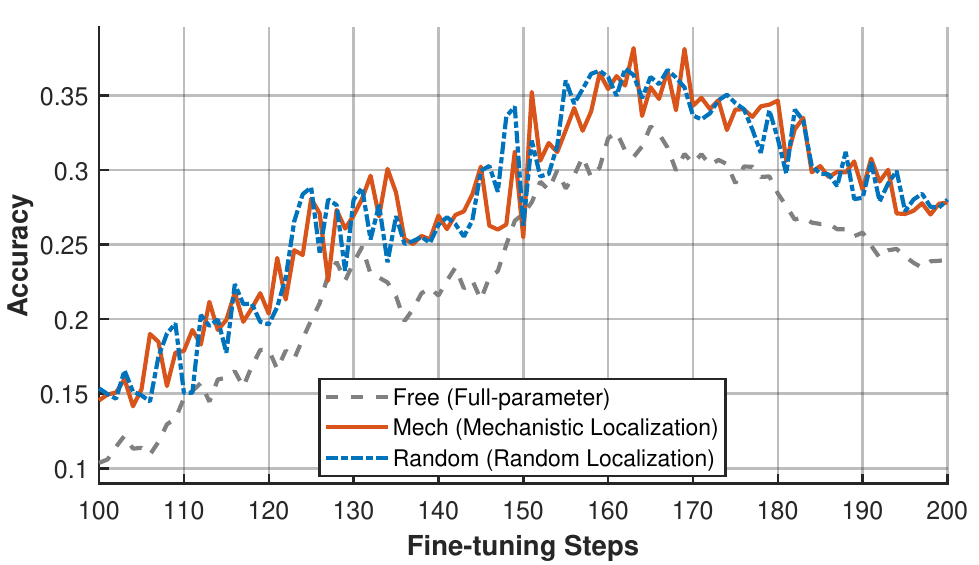}}
    \subfigure[CC of Gender]{
    \includegraphics[width=0.31\linewidth]{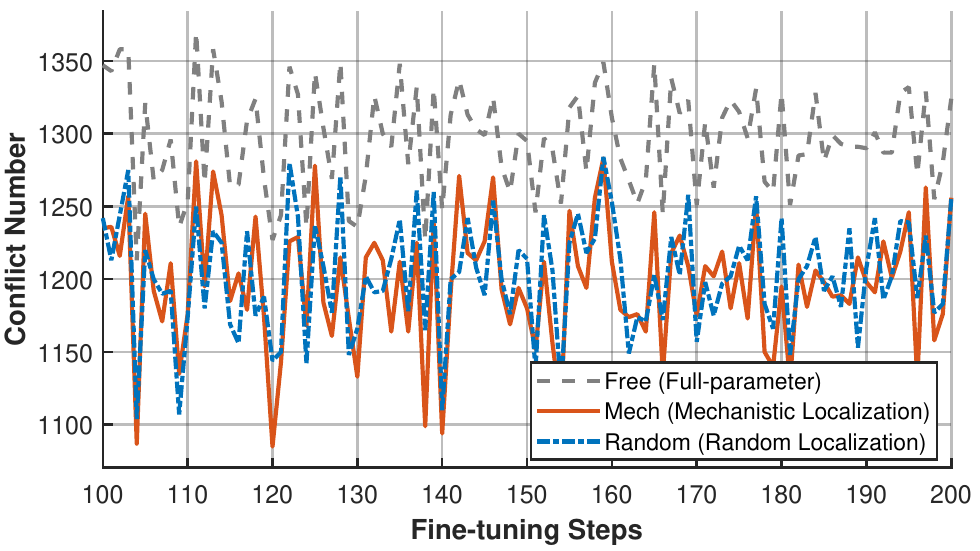}}

    \subfigure[T-ACC of Winogrande]{
    \includegraphics[width=0.31\linewidth]{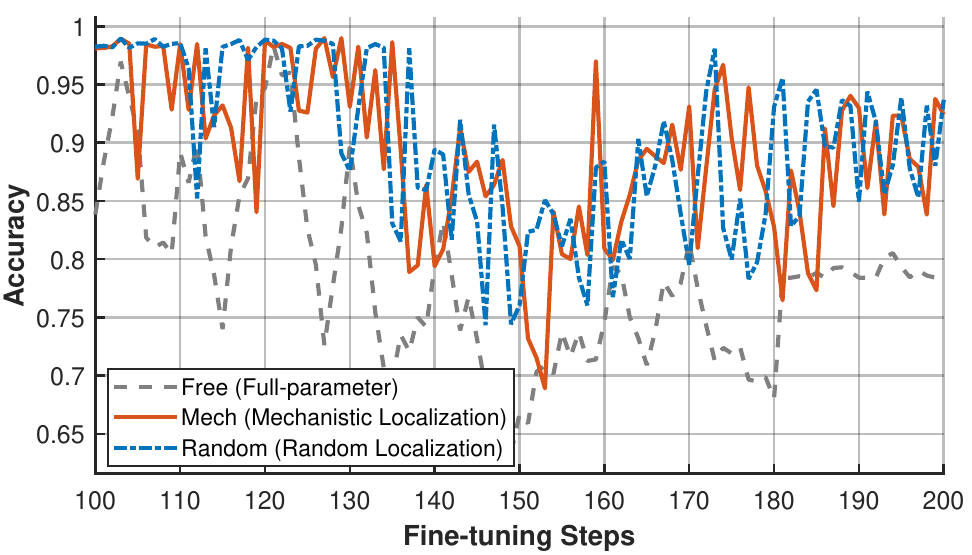}}
    \subfigure[P-Acc of Winogrande]{
    \includegraphics[width=0.31\linewidth]{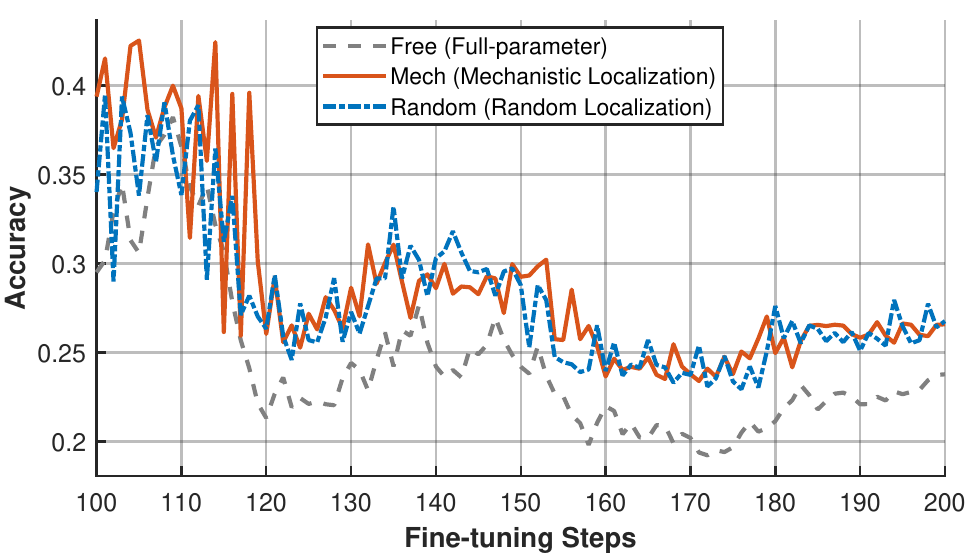}}
    \subfigure[CC of Winogrande]{
    \includegraphics[width=0.31\linewidth]{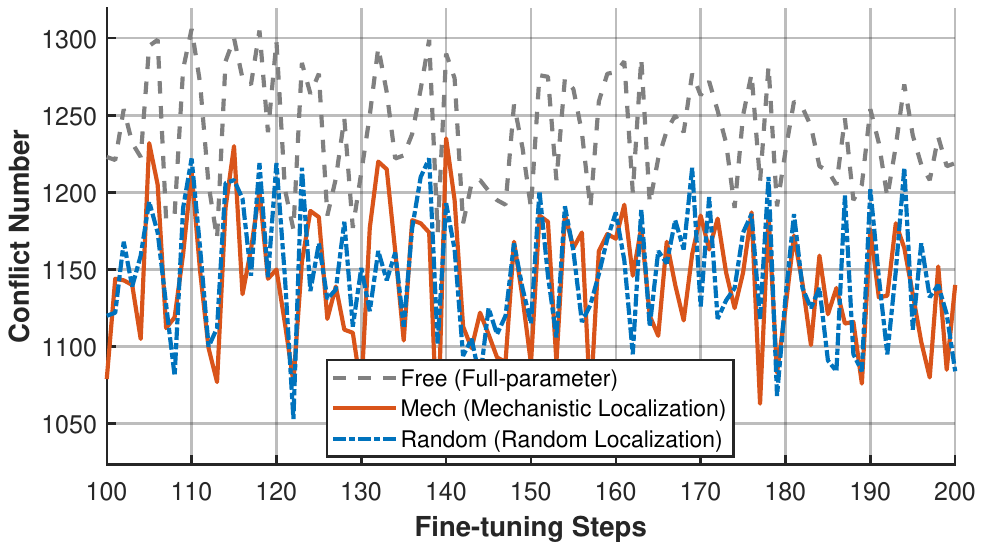}}

    \subfigure[T-ACC of SST2]{
    \includegraphics[width=0.31\linewidth]{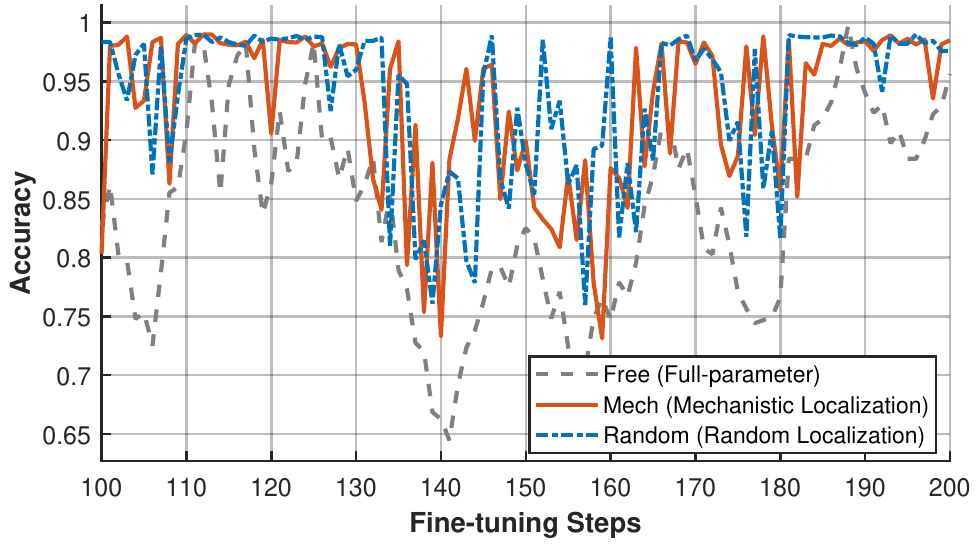}}
    \subfigure[P-Acc of SST-2]{
    \includegraphics[width=0.31\linewidth]{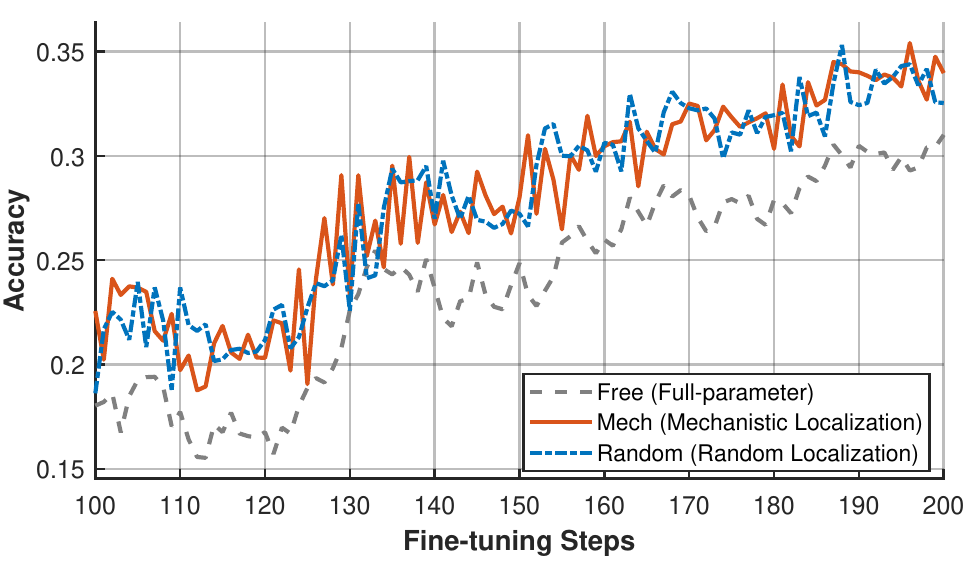}}
    \subfigure[CC of SST-2]{
    \includegraphics[width=0.31\linewidth]{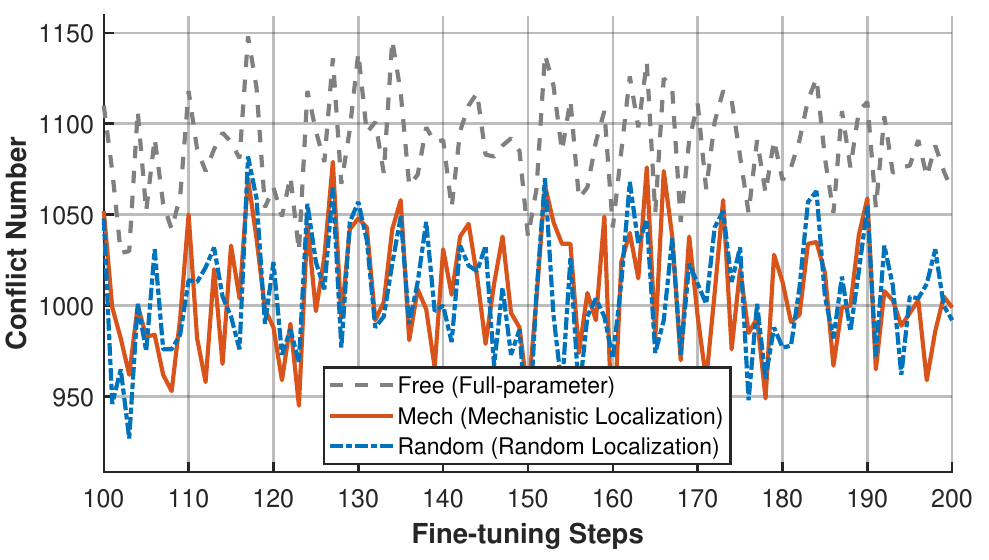}}
    \caption{Target Task Accuracy (T-Acc), Pervasiveness Task Accuracy (P-Acc), and Circuit Conflict (CC) of Bool, Gender, Winogrande, and SST-2 Task with localization. }
    \label{fig:app_localization_others}
\end{figure*}

Figure~\ref{fig:app_localization_others} details the evolutionary trajectories of the remaining four target tasks across the Accuracy and Circuit Conflict metrics. Corroborating the findings from the Arithmetic task, the localization methods consistently exhibit superior performance relative to the \textit{Free} evolution baseline.

\subsection{Impact of Localization on Circuit Distance and Stability}
\label{subsec:app_f2_cd_cs}

\begin{figure*}
    \centering
     \subfigure[$CD$ of Arithmetic]{
    \includegraphics[width=0.45\linewidth]{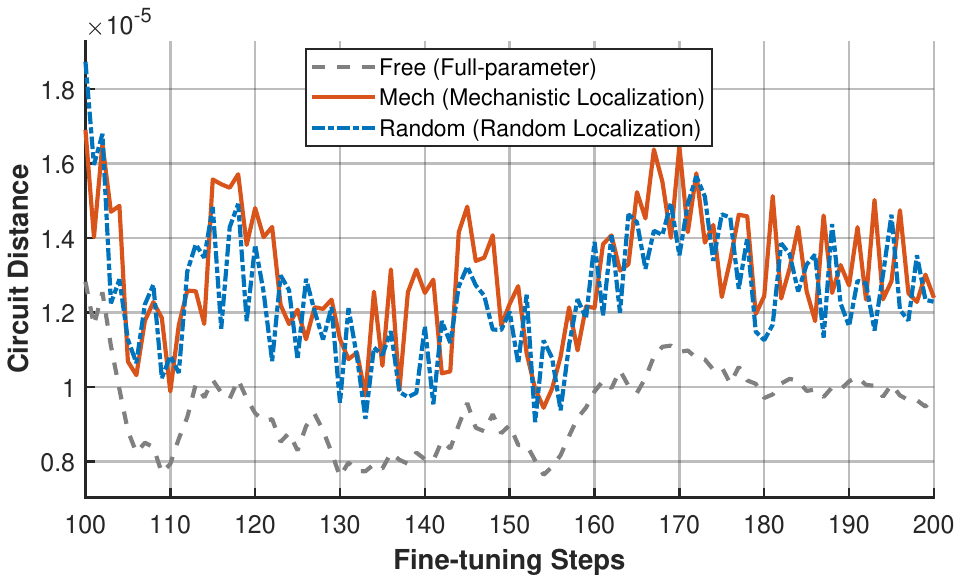}}
    \subfigure[$CS$ of Arithmetic]{
    \includegraphics[width=0.45\linewidth]{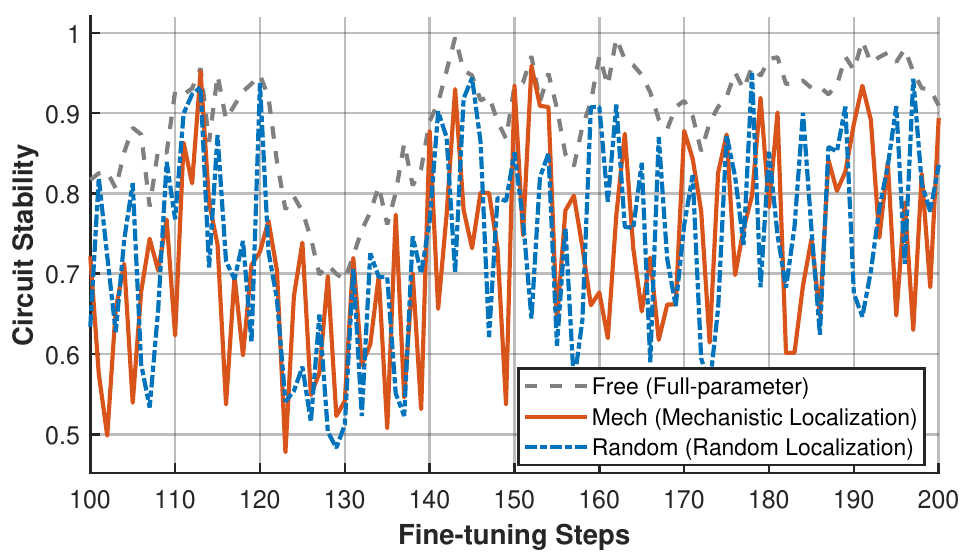}}
    
    \subfigure[$CD$ of Bool]{
    \includegraphics[width=0.45\linewidth]{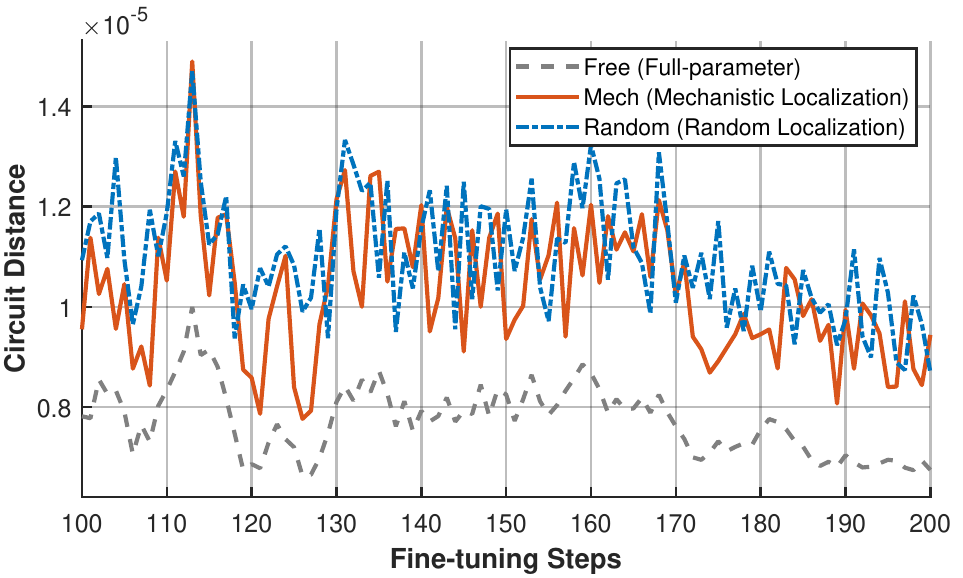}}
    \subfigure[$CS$ of Bool]{
    \includegraphics[width=0.45\linewidth]{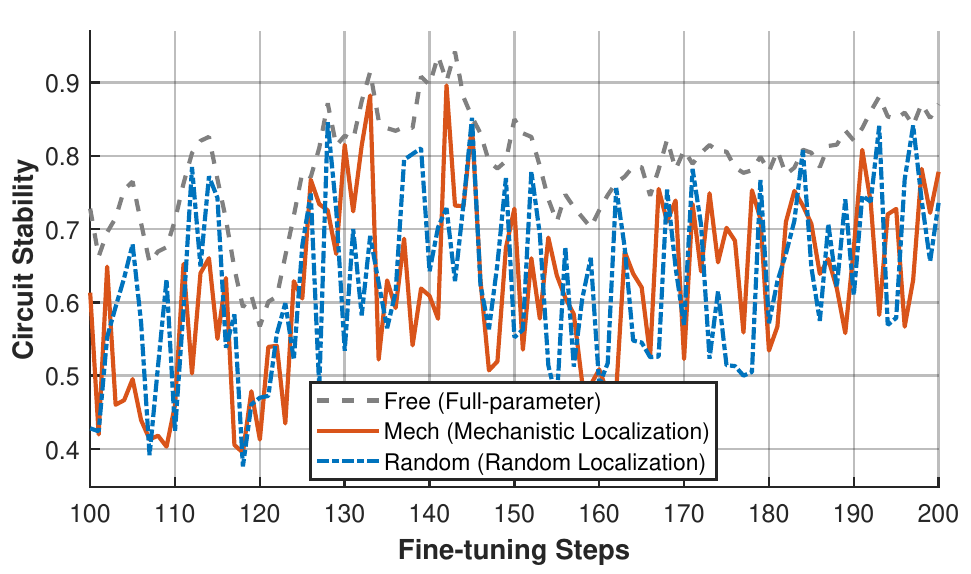}}

    \subfigure[$CD$ of Gender]{
    \includegraphics[width=0.45\linewidth]{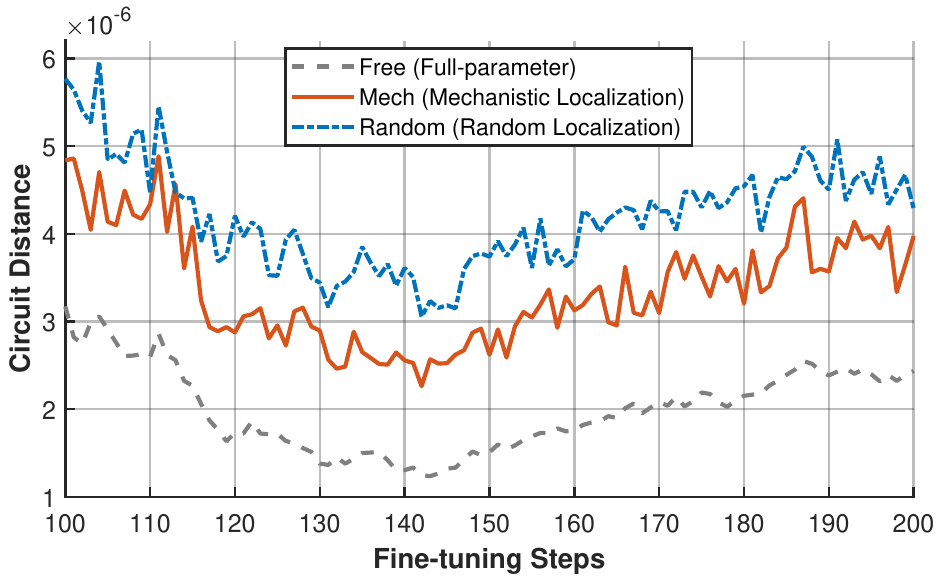}}
    \subfigure[$CS$ of Gender]{
    \includegraphics[width=0.45\linewidth]{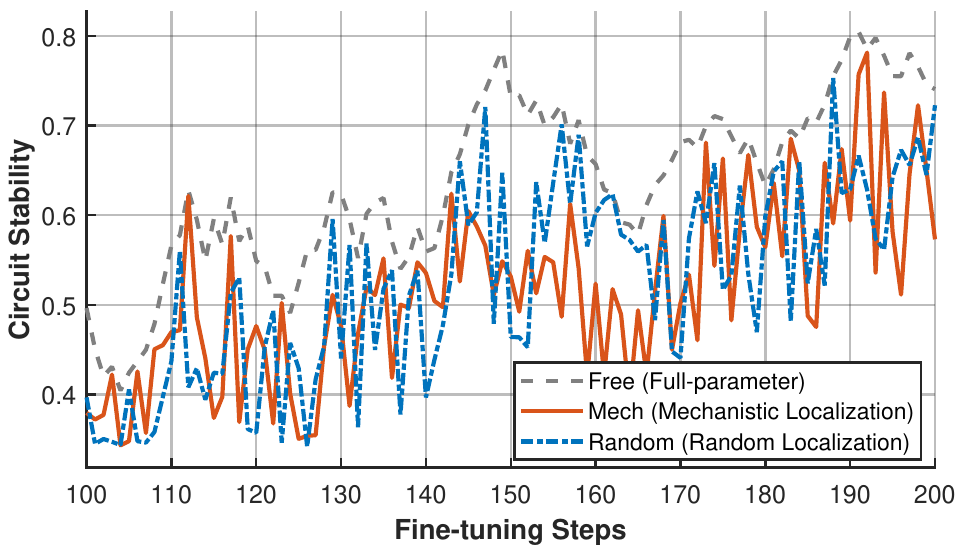}}

    \subfigure[$CD$ of Winogrande]{
    \includegraphics[width=0.45\linewidth]{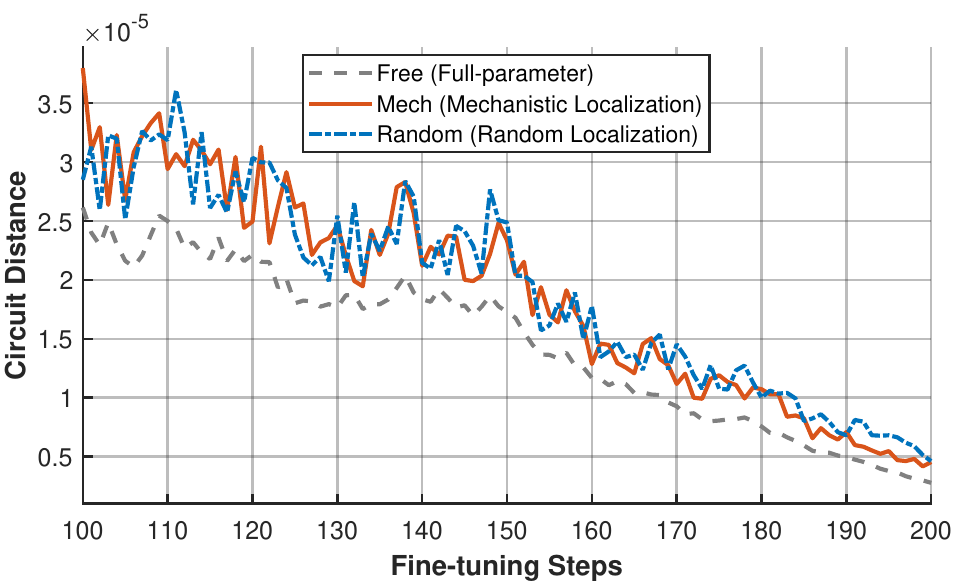}}
    \subfigure[$CS$ of Winogrande]{
    \includegraphics[width=0.45\linewidth]{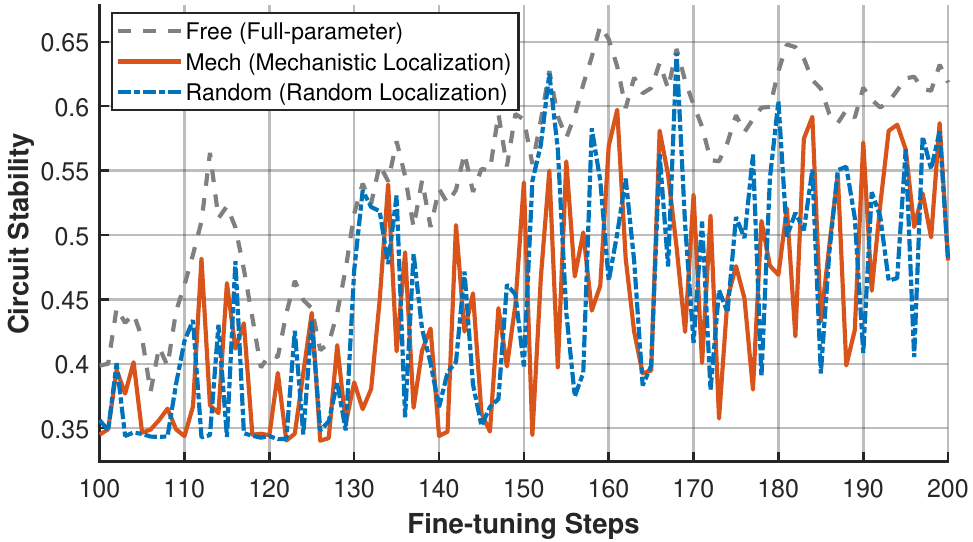}}

    \subfigure[$CD$ of SST2]{
    \includegraphics[width=0.45\linewidth]{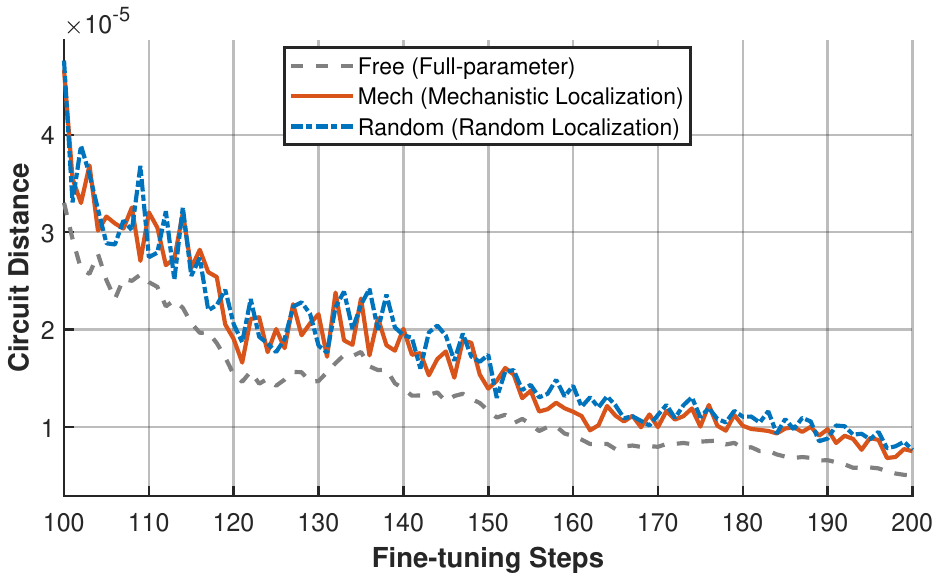}}
    \subfigure[$CS$ of SST-2]{
    \includegraphics[width=0.45\linewidth]{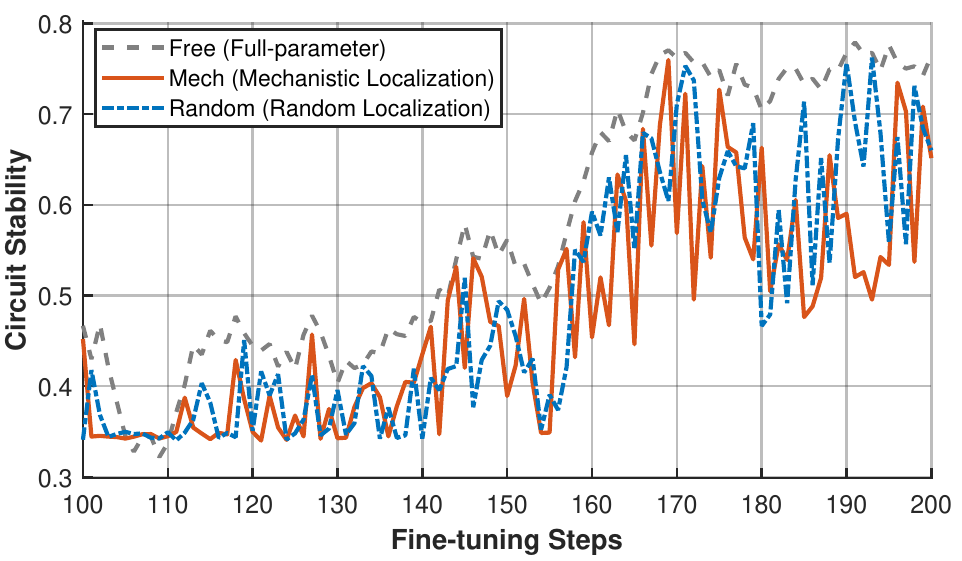}}
    
    \caption{Circuit Distance ($CD$) and Circuit Stability ($CS$) of Arithmetic, Bool, Gender, Winogrande, and SST-2 Task with localization. }
    \label{fig:app_localization_cd_cs}
\end{figure*}

Figure~\ref{fig:app_localization_cd_cs} illustrates the performance of the five target tasks concerning Circuit Distance ($CD$) and Circuit Stability ($CS$). A pronounced observation is that localization methods precipitate substantially greater component migration compared to free evolution. This strongly implies that the components pinpointed by localization do not constitute the genuinely optimal subset required for adaptation; paradoxically, this artificial restriction renders the evolutionary trajectory even more ``incorrect'' than if no localization were applied. Concurrently, the sharp decline in $CS$ demonstrates that it becomes markedly more arduous to effectively update internal information within the remaining unfrozen components. Collectively, these phenomena substantiate that current Mechanistic Localization fails to isolate the truly critical components and inadvertently freezes components essential for natural adaptation. It is precisely this misallocation that induces the chaotic volatility observed in the individual circuit metrics.

\subsection{Impact of Circuit Scale on Localization Efficacy}
\label{subsec:app_f3_circuit_scale}

To further investigate the limitations of static localization, we conducted a systematic interpolation study on the circuit scale. We evaluated task performance (T-Acc and P-Acc), Circuit Distance ($CD$), Circuit Stability ($CS$), and Circuit Conflict ($CC$) across varying circuit capacities: $500$, $800$, $1,000$, $1,500$, $2,000$, $2,500$, $3,000$, $3,500$, and $4,000$ (corresponding to the total number of unfrozen critical components).

As demonstrated in Table~\ref{tab:app_circuit_scale}, optimal performance across all metrics is achieved when the circuit scale expands to approximately $3,000$--$3,500$ components. At this capacity, the localized subset is sufficiently expansive to encompass those latent components that---while deemed ``marginally important'' under the current parameter state---prove highly critical for future parameter updates. This highlights a fundamental trade-off inherent in determining circuit scale: an overly restrictive scale reliably isolates components critical to the \textit{current} state but inadvertently discards latent components vital for \textit{future} evolution; conversely, an excessively permissive scale severely compromises the very ``interpretability'' the mechanism seeks to provide. 

\begin{table}[ht]
  \caption{Performance of different circuit scale in Arithmetic task}
  \label{tab:app_circuit_scale}
  \centering
  \resizebox{\textwidth}{!}{
  \begin{tabular}{lllllllllll}
    \toprule
  Localization& Metric  &$500$   & $800$   & $1,000$   & $1,500$   & $2,000$   & $2,500$   & $3,000$  & $3,500$  &$4,000$\\
    \midrule
    \multirow{5}{*}{Mech}& T-Acc &0.75  &0.76     &0.73       &0.71       &0.73       &0.73& \textbf{0.75}& \textbf{0.75}&0.72\\
    &P-Acc& 0.26& 0.25& 0.24& 0.25&0.26&0.24&\textbf{0.26}&\textbf{0.27}&0.25\\
    &$CD$($1*10^{-5}$)&1.38&1.24&1.28&\textbf{1.26}&1.27&\textbf{1.33}&\textbf{1.35}&\textbf{1.32}&1.29\\
    &$CS$&0.84&\textbf{0.86}&0.81&\textbf{0.75}&0.88&0.75&\textbf{0.70}&\textbf{0.65}&\textbf{0.61}\\
    &$CC$&1105&1247&\textbf{1067}&1134&1212&\textbf{1069}&\textbf{1109}&\textbf{1142}&\textbf{1129}\\
\hline
    \multirow{5}{*}{Random}& T-Acc&0.75  &0.75     &0.74       &0.72       &0.73       &0.72& 0.73& 0.74&0.73\\
     &P-Acc& 0.25& 0.25& 0.25& 0.24&0.27&0.25&0.24&0.25&0.25\\
     &$CD$($1*10^{-5}$)&1.33&1.21&1.26&1.28&1.27&1.38&1.41&1.36&1.25\\
     &$CS$&0.85&0.82&0.86&0.74&0.84&0.77&0.60&0.58&0.53\\
     &$CC$&1006&1212&1075&1123&1201&1077&1215&1194&1168\\
    \bottomrule
  \end{tabular}}
\end{table}

Crucially, however, at this optimal scale of $3,000$--$3,500$ components, the localized circuit encompasses nearly $80\%$ of the language model's total target components. Consequently, any assertion regarding the inherent ``effectiveness of localization'' becomes exceedingly tenuous at this juncture, as the tuning process fundamentally regresses toward standard full-parameter SFT.

Additionally, we evaluated the comparative efficacy of the three distinct circuit extraction methodologies outlined in Appendix~\ref{supplogicalcircuit}. Table~\ref{tab:circuit_extraction_comparison} delineates the performance disparities among ACDC, EAP, and EdgePruning. It is evident that the EAP approach markedly outperforms the other two alternatives. Synthesizing this empirical finding with our analysis of \textbf{RQ2-c} in Section~\ref{subsec:localization_evolution}, a compelling explanation emerges: EAP's reliance on gradient-based estimation aligns inherently with the trajectory of gradient descent. Consequently, this characteristic renders the EAP-derived circuits substantially more compatible with, and adaptive to, future dynamic parameter updates.

\begin{table}[ht]
  \caption{Performance of different circuit methodologies}
  \label{tab:circuit_extraction_comparison}
  \centering
  \resizebox{0.8\textwidth}{!}{
  \begin{tabular}{lllllll}
    \toprule
  Dataset&Circuit& T-Acc  &P-Acc&$CD$($1*10^{-5}$)&$CS$&$CC$\\
    \midrule
   \multirow{3}{*}{Arithmetic}& EAP&0.73&0.26&1.27&0.88&1212\\
   &ACDC&0.71&0.22&1.42&0.87&1154\\
   &EdgePruning&0.74&0.25&1.55&0.87&1205\\

   \multirow{3}{*}{Gender}&EAP& 0.91&0.28&3.27&0.68&1258\\
   &ACDC&0.88&0.25&4.66&0.75&1239\\
   &EdgePruning&0.84&0.30&3.69&0.61&1349\\
    \bottomrule
  \end{tabular}}
\end{table}

\section{Validation in Single-Objective SFT}
\label{appendix:single_objective}

In Section 4.2.2, we established a critical conclusion: in tasks governed by MLP-dominated circuits, the circuits are inherently more resistant to migration. Consequently, Mechanistic Localization coincidentally retains its guiding significance for future parameter updates, yielding performance metrics significantly superior to both Random Localization and free evolution. Conversely, for tasks governed by Attention-dominated circuits, intense component migration precipitates massive structural discrepancies between the circuit under the current parameter state and circuits under future states. Thus, Mechanistic Localization fails to provide predictive guidance for future parameters, resulting in performance virtually indistinguishable from Random Localization.

To rigorously validate this conclusion, we eliminated the pervasiveness task during the SFT of the Mistral-7B model. This allowed us to observe whether the isolated target task exhibits similar evolutionary patterns under pure single-objective optimization.

\begin{table}[ht]
  \caption{Performance of WMDP-Bio dataset without pervasiveness task}
  \label{tab:app_single_objective_WMDP}
  \centering
  \resizebox{\textwidth}{!}{
  \begin{tabular}{lllllllllllll}
    \toprule
  Metric& Localization  &Epoch 0   &Epoch 1 &Epoch 2 &Epoch 3 &Epoch 4 &Epoch 5 &Epoch 6 &Epoch 7 &Epoch 8 &Epoch 9 &Epoch 10\\
    \midrule
    \multirow{3}{*}{ACC }&Free&0.36&0.01&0.00&0.00&0.00&0.15&0.26&0.45&0.49&0.59&0.72\\
                       &Random&0.36&0.00&0.00&0.04&0.26&0.19&0.35&0.55&0.68&0.72&0.75\\
                         &Mech&0.36&0.00&0.00&0.00&0.15&0.38&0.67&0.59&0.77&0.82&0.85\\
    \hline
    \multirow{3}{*}{$CD$}&Free&0&0.95&1.12&1.23&1.15&1.08&1.02&0.95&0.98&0.94&0.93\\
                       &Random&0&1.34&1.44&1.52&1.59&1.51&1.55&1.35&1.39&1.25&1.21\\
                         &Mech&0&1.59&1.66&1.54&1.42&1.39&1.21&1.14&1.19&1.11&1.11\\
    \hline
    \multirow{3}{*}{$CS$}&Free&0.78&0.35&0.29&0.21&0.39&0.44&0.64&0.58&0.66&0.75&0.84\\
                       &Random&0.78&0.29&0.15&0.34&0.46&0.42&0.51&0.67&0.79&0.75&0.89\\
                         &Mech&0.78&0.33&0.29&0.27&0.62&0.35&0.55&0.61&0.79&0.85&0.94\\
    \hline
    \multirow{3}{*}{$CC$}&Free& 982&1057&1132&1164&1192&1167&1158&1138&1146&1142&1084\\
                       &Random& 982&1145&1231&1214&1195&1194&1163&1147&1195&1175&1181\\
                         &Mech& 982&1129&1178&1235&1207&1151&1129&1166&1157&1121&1096\\
    \bottomrule
  \end{tabular}}
\end{table}

\begin{table}[ht]
  \caption{Performance of Induction dataset without pervasiveness task}
  \label{tab:app_single_objective_Induction}
  \centering
  \resizebox{\textwidth}{!}{
  \begin{tabular}{lllllllllllll}
    \toprule
  Metric& Localization  &Epoch 0   &Epoch 1 &Epoch 2 &Epoch 3 &Epoch 4 &Epoch 5 &Epoch 6 &Epoch 7 &Epoch 8 &Epoch 9 &Epoch 10\\
    \midrule
    \multirow{3}{*}{ACC }&Free&0.45&0.05&0.00&0.00&0.00&0.29&0.64&0.38&0.79&0.66&0.88\\
                       &Random&0.45&0.00&0.00&0.00&0.29&0.67&0.49&0.33&0.81&0.67&0.92\\
                         &Mech&0.45&0.00&0.00&0.00&0.00&0.54&0.39&0.67&0.81&0.74&0.89\\
    \hline
    \multirow{3}{*}{$CD$}&Free&0&1.25&1.33&1.65&1.41&1.58&1.51&1.46&1.48&1.34&1.35\\
                       &Random&0&1.67&1.88&1.94&1.92&1.87&1.87&1.83&1.75&1.79&1.69\\
                         &Mech&0&1.69&1.92&1.85&1.98&1.75&1.92&1.96&1.88&1.85&1.72\\
    \hline
    \multirow{3}{*}{$CS$}&Free&0.63&0.23&0.16&0.31&0.25&0.41&0.48&0.72&0.79&0.85&0.89\\
                       &Random&0.63&0.29&0.38&0.19&0.44&0.51&0.67&0.82&0.85&0.89&0.95\\
                         &Mech&0.63&0.39&0.34&0.35&0.61&0.59&0.76&0.82&0.89&0.93&0.96\\
    \hline
    \multirow{3}{*}{$CC$}&Free&1057&1244&1341&1319&1267&1294&1288&1276&1255&1219&1284\\
                       &Random&1057&1209&1185&1167&1139&1194&1205&1185&1157&1167&1139\\
                         &Mech&1057&1257&1288&1234&1209&1184&1193&1172&1164&1153&1149\\
    \bottomrule
  \end{tabular}}
\end{table}

Table~\ref{tab:app_single_objective_WMDP} and~\ref{tab:app_single_objective_Induction} unequivocally demonstrates that when WMDP-Bio and Induction are optimized independently as isolated target tasks, the following conclusions persistently hold true:
\begin{enumerate}
    \item \textbf{MLP-dominated circuits} are significantly less prone to migration. Therefore, Mechanistic Localization for these knowledge-centric tasks genuinely imparts meaningful guidance for future parameter updates.
    \item \textbf{Attention-dominated circuits} are highly susceptible to migration, leading to profound structural discrepancies across different parameter states. Consequently, Mechanistic Localization for these skill-centric tasks suffers from severe temporal latency, rendering it ineffective for guiding dynamic updates.
\end{enumerate}

\section{Extended Analysis of Future Mechanistic Localization and Methodological Comparisons}
\label{suppmorefuture}

\begin{figure*}
    \centering
    \subfigure[Circuit Distance]{
    \includegraphics[width=0.31\linewidth]{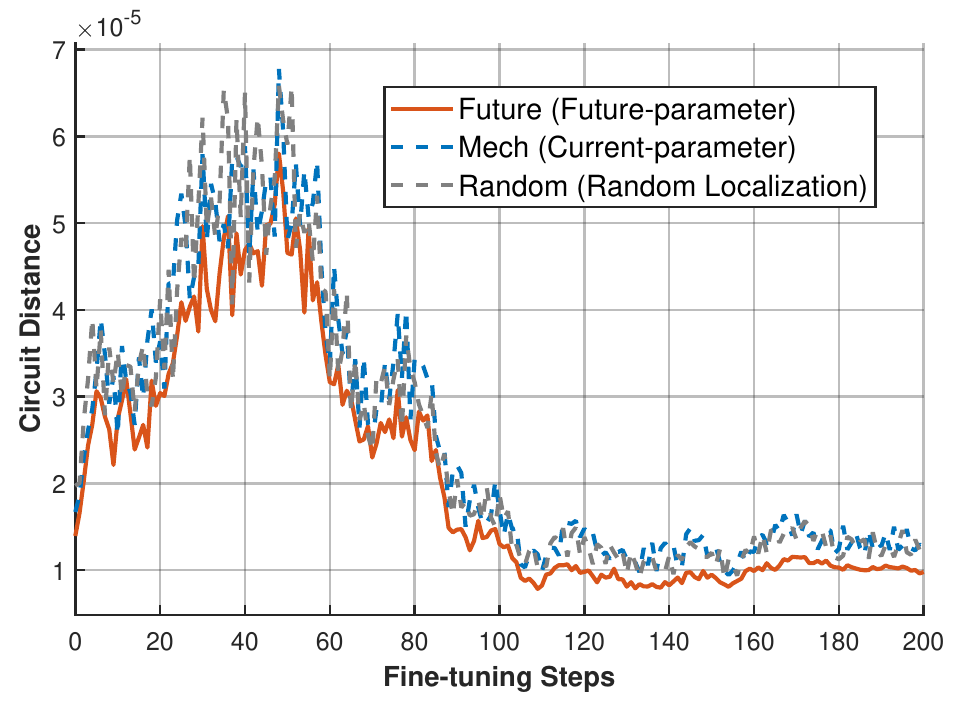}}
    \subfigure[Circuit Stability]{
    \includegraphics[width=0.31\linewidth]{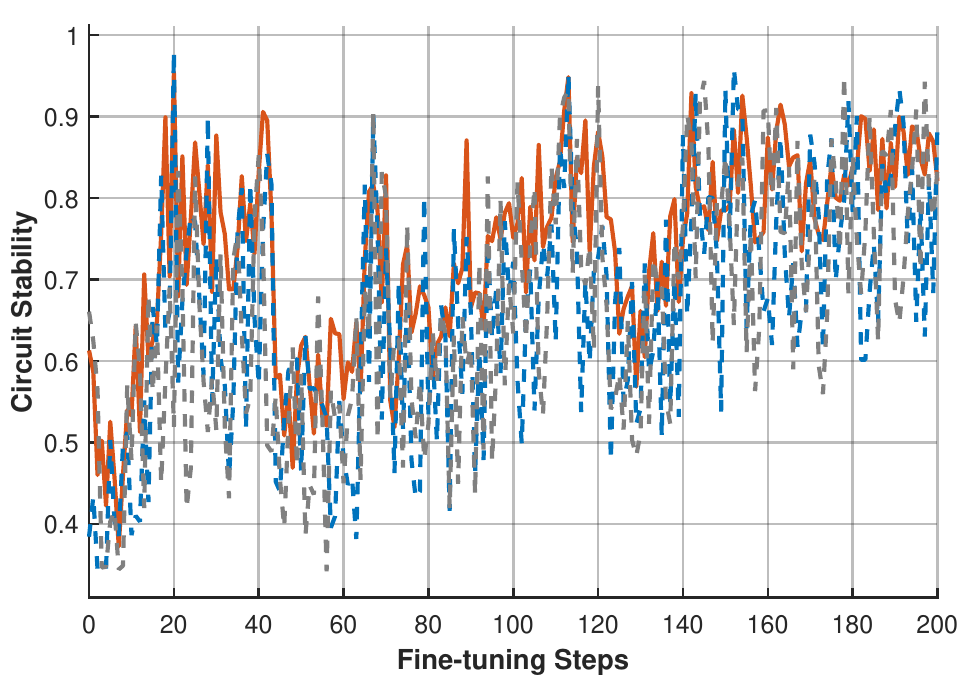}}
    \subfigure[Circuit Conflict]{
    \includegraphics[width=0.31\linewidth]{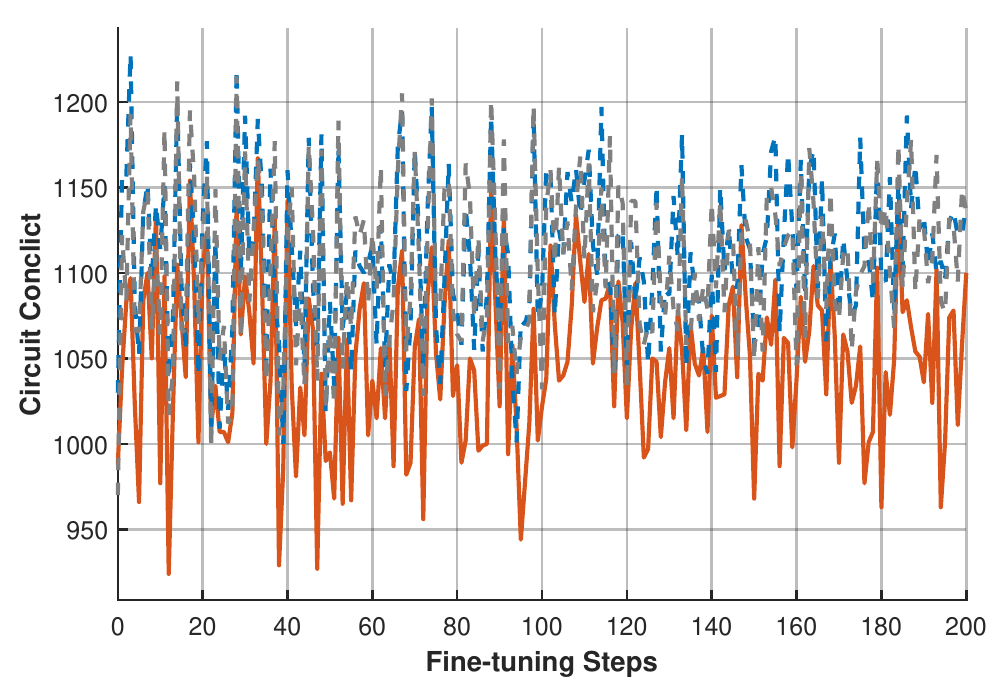}}
    \caption{Line plots of Future Mechanistic Localization }
    \label{fig:app_future_mechanistic_metrics}
\end{figure*}

First, we present the comprehensive circuit metrics---Circuit Distance ($CD$), Circuit Stability ($CS$), and Circuit Conflict ($CC$)---for the Arithmetic task under the \textit{Future Mechanistic} localization paradigm, as illustrated in Figure~\ref{fig:app_future_mechanistic_metrics}. These detailed metrics further substantiate the profound superiority and stability of the \textit{Future Mechanistic} approach over static baseline methods.

Subsequently, we expand our comparative analysis beyond standard circuit discovery to encompass other predominant Mechanistic Localization methodologies. Specifically, we evaluate gradient-based methods (e.g., WAGLE and DEPN), intervention-based methods (e.g., MEMIT and CLUE), and strictly circuit-centric methods (e.g., CLUE-EAP and CLUE-EdgePruning). 

As demonstrated in Table~\ref{tab:app_localization_methods_comparison}, gradient-associated methodologies consistently yield the optimal $CD$ and $CS$ metrics among all evaluated techniques. Notably, this superior performance persists even for methods that do not rely on standard gradient descent fine-tuning for parameter updates (e.g., MEMIT, which employs closed-form vector editing). 
The introduction of these strategies of Localization are as follows:

\textbf{DEPN}~\citep{wu2023depn} (Detect and Edit Privacy Neurons) is a framework designed to safeguard against privacy leakage in pretrained language models by localizing and editing specific neurons. The method's core localization component is a novel privacy neuron detector that uses a gradient-based attribution technique. This detector computes a privacy attribution score for each neuron to quantify its contribution to the model's leakage of private information. This is achieved by calculating the cumulative gradient of the output probability with respect to the neuron's activation value, as the activation is gradually changed from zero to its original value. 

\textbf{WAGLE}~\citep{jia2024wagle} (Weight Attribution-guided LLM Unlearning Framework) is a framework that pinpoints the most influential weights for unlearning through a strategic weight attribution method. The method frames the weight attribution problem as a bi-level optimization (BLO) problem, which allows it to balance unlearning efficacy with utility preservation. The core of the localization process is the derivation of a closed-form attribution score for each weight, calculated using the implicit gradient from the BLO problem. This score's value is determined by combining the gradients from both the forget loss and the retain loss.

\textbf{MEMIT}~\citep{patilcan} ((Mass-Editing Memory in a Transformer)) addresses the deletion of factual information by causal tracing, a denoising-based intervention method. This approach relies on the assumption that knowledge is stored in specific, localized components of the network, and can be identified via causal mediation. 

\textbf{KN}~\citep{dai2022knowledge} (Knowledge Neurons) introduces the concept of knowledge neurons to investigate how factual knowledge is stored in pretrained Transformers. As an intervention-based method, it views feed-forward network (FFN) modules as key-value memories. The method utilizes a knowledge attribution technique based on integrated gradients to evaluate the contribution of each neuron to knowledge predictions. By identifying and manipulating (e.g., suppressing or amplifying) these specific neurons, KN demonstrates that interventions on localized neurons can explicitly affect knowledge expression and edit factual knowledge within the model without the need for SFT.

\textbf{CLUE}~\citep{chen2025clue} (Conflict-guided Localization for LLM Unlearning Framework) is a circuit-based localization framework designed to improve the precision of LLM unlearning. It leverages mechanistic interpretability to discover logical circuits corresponding to the forget set and the retain set. CLUE transforms these circuits into Conjunctive Normal Form (CNF) and uses a Boolean satisfiability solver to disentangle the intertwined nodes into three distinct categories: forget nodes, retain nodes, and conflict nodes. By pinpointing the specific function of each node, CLUE enables targeted SFT strategies that significantly enhance forget efficacy while preserving retain utility, avoiding the pitfalls of applying uniform interventions on entangled nodes.

This empirical evidence strongly implies that gradient-based techniques inherently possess greater ``foresight'' during the parameter update process; they are capable of prospectively identifying a crucial subset of the critical components that will ultimately govern the future parameter state.

\begin{table}[ht]
  \caption{Performance of different Localization Strategies on Arithmetic Dataset}
  \label{tab:app_localization_methods_comparison}
  \centering
  \resizebox{0.9\textwidth}{!}{
  \begin{tabular}{lllllllllllll}
    \toprule
  Categories& Localization  &Updating&$CD$&$CS$&$CC$&T-Acc&P-Acc\\
    \midrule
    \multirow{2}{*}{Gradient}&WAGLE&Fine-Tuning&1.31&0.89&1157&0.74&0.25\\
   & DEPN&Fine-Tuning&1.29&0.85&1137&0.72&0.26\\
    \hline
    \multirow{2}{*}{Intervention}&MEMIT&Editing&1.64&0.74&1344&0.71&0.22\\
    &KN&Editing&1.68&0.72&1315&0.72&0.23\\
    \hline
    \multirow{2}{*}{Circuit}&CLUE (EdgePruning)&Fine-Tuning&1.55&0.87&1205&0.74&0.25\\
    &CLUE (EAP)&Fine-Tuning&1.27&0.88&1212&0.73&0.26\\
    
    \bottomrule
  \end{tabular}}
\end{table}

\section{Limitations}\label{supplimitation}

While this paper derives a series of insightful conclusions by observing transformer circuits throughout the Supervised Fine-Tuning (SFT) process, several limitations remain to be acknowledged:

\begin{enumerate}
    \item \textbf{Inherent Limitations of Circuit Discovery:} The process of circuit discovery itself is notoriously difficult to scale to exceptionally large LLMs and imposes stringent requirements on data quality. Consequently, this computational bottleneck precludes further analysis under massive data and model scaling scenarios, thereby restricting the direct application of our analytical framework in certain real-world, large-scale deployments.
    \item \textbf{Coupling of Localization and Parameter Update Mechanisms:} Many contemporary Mechanistic Localization methodologies introduce bespoke parameter update techniques paired with their localization strategies; the effects of these two components are rarely strictly independent. Although employing standard SFT as our observational baseline allows us to capture universally applicable and dynamic evolutionary trends, integrating our framework with these specialized, coupled update methods could potentially unveil more granular and nuanced patterns that remain unexplored in this work.
\end{enumerate}
%%%%%%%%%%%%%%%%%%%%%%%%%%%%%%%%%%%%%%%%%%%%%%%%%%%%%%%%%%%%

\end{document}